%% file: main.tex
\documentclass{article}

\usepackage{iclr2027_conference,times}

\usepackage[utf8]{inputenc}
\usepackage[T1]{fontenc}
\usepackage{hyperref}
\usepackage{url}
\usepackage{amsfonts}
\usepackage{amsmath}
\usepackage{nicefrac}
\usepackage{xcolor}
\usepackage{colortbl}
\definecolor{collabphase}{HTML}{3A8F2A}
\definecolor{noncollabphase}{HTML}{1B4F9B}
\newcommand{\collabphase}[1]{\textcolor{collabphase}{\boldmath #1}}
\newcommand{\noncollabphase}[1]{\textcolor{noncollabphase}{\boldmath #1}}
\usepackage{graphicx}
\usepackage{wrapfig}
\usepackage{adjustbox}
\usepackage{multirow}
\usepackage{pifont}
\usepackage{xspace}
\usepackage{capt-of}
\usepackage{placeins}
\usepackage{booktabs}
\usepackage{tabularx}
\usepackage{longtable}
\usepackage{etoolbox}
\makeatletter
\patchcmd{\LT@makecaption}{\reset@font}{\reset@font\normalsize}{}%
  {\PackageWarningNoLine{cohub}{longtable caption size patch failed}}
\patchcmd{\LT@makecaption}{\endgraf\vskip\baselineskip}{\endgraf\vskip\dimexpr\belowcaptionskip+\lineskip\relax}{}%
  {\PackageWarningNoLine{cohub}{longtable caption skip patch failed}}
\makeatother
\usepackage{array}
\usepackage{makecell}
\usepackage{enumitem}
\usepackage{pgfplots}  
\pgfplotsset{compat=1.18}
\usepgfplotslibrary{groupplots}
\usepackage[scaled=0.82]{beramono}

\newcolumntype{C}{>{\centering\arraybackslash}X}
\newcommand{\bench}{\textsc{CoHuB}\xspace}
\newcommand{\yes}{\ding{51}}
\newcommand{\no}{\ding{55}}

\newcommand{\na}{--}

\newlength{\taskstripw}
\newlength{\taskframew}
\newlength{\taskimgsep}

\newcommand{\taskcropside}{0.21875}

\newcommand{\taskimg}[1]{%
  \adjustbox{trim={\taskcropside\width} 0 {\taskcropside\width} 0,clip,%
             width=\taskframew}{\includegraphics{#1}}%
}

\newcommand{\taskstrip}[4]{%
  \taskimg{#1/#2}\hspace{\taskimgsep}%
  \taskimg{#1/#3}\hspace{\taskimgsep}%
  \taskimg{#1/#4}%
}

\newcommand{\taskcell}[2]{%
  \begin{minipage}[t]{\taskstripw}
    \centering
    #2\\[2pt]
    {\small\textbf{#1}}
  \end{minipage}%
}

\newlength{\failstripw}
\newlength{\failimgsep}
\newlength{\failimgw}
\newcommand{\failcropside}{0}
\newcommand{\failimg}[1]{%
  \IfFileExists{#1}{%
    \adjustbox{trim={\failcropside\width} 0 {\failcropside\width} 0,clip,%
               width=\failimgw}{\includegraphics{#1}}%
  }{%
    \fcolorbox{gray}{white}{\begin{minipage}[b][\dimexpr0.75\failimgw-2\fboxsep-2\fboxrule\relax][c]{\dimexpr\failimgw-2\fboxsep-2\fboxrule\relax}%
      \centering\scriptsize\textcolor{gray}{render pending}\end{minipage}}%
  }%
}
\newcommand{\failcell}[3]{%
  \begin{minipage}[t]{\failstripw}
    \centering
    \failimg{#2}\hspace{\failimgsep}\failimg{#3}\\[2pt]
    {\small\textbf{#1}}
  \end{minipage}%
}

\newcommand{\failsetlengths}{%
  \setlength{\failstripw}{0.485\linewidth}%
  \setlength{\failimgsep}{0.01\linewidth}%
  \setlength{\failimgw}{\dimexpr(\failstripw-\failimgsep)/2\relax}%
}

\title{\bench: A Simulation Benchmark for Multi-Humanoid Collaboration}

\author{%
  Hyunjin Park$^{1*\dagger}$\quad Jebeom Chae$^{1*}$\quad Minwoo Park$^{1*}$\quad Sunghyun Park$^{1}$\quad Hanjun Yoo$^{2}$ \\
  \textbf{Seoyeon Choi$^{3}$\quad Soochul Yoo$^{1}$\quad Joohwan Seo$^{3}$\quad Sarmad Idrees$^{1}$\quad Jae-Sang Hyun$^{1}$} \\
  \textbf{Jongmin Lee$^{1}$\quad Roberto Horowitz$^{3}$\quad Youngwoon Lee$^{4}$\quad Jongeun Choi$^{1\ddagger}$} \\[6pt]
  $^{1}$Yonsei University\quad $^{2}$Gwangju Institute of Science and Technology \\
  $^{3}$University of California, Berkeley\quad $^{4}$Seoul National University \\[4pt]
  {\small $^{*}$Co-first authors\quad $^{\dagger}$Project lead\quad $^{\ddagger}$Corresponding author}
}

\iclrfinalcopy
\begin{document}

\maketitle
\lhead{Preprint}

\begin{abstract}
Many physical tasks in human environments require collaboration, from assisting a partner to jointly manipulating an object.
Yet, existing humanoid benchmarks largely focus on single-humanoid skills and lack evaluation of multi-humanoid collaboration under egocentric visual observations.
We introduce \mbox{\bench} (\textbf{Co}llaborative Multi-\textbf{Hu}manoid \textbf{B}enchmark), a simulation benchmark for multi-humanoid collaboration under egocentric visual observations.
\bench provides 10 tasks---eight with two humanoids and two with three humanoids---spanning diverse collaboration patterns. We also provide synchronized demonstrations collected through a multi-operator VR teleoperation pipeline, in which each operator controls one humanoid from its egocentric view.
Experiments with representative visuomotor policies reveal substantial challenges across different forms of coordinated perception and control. \bench provides a foundation for developing and evaluating multi-humanoid collaboration policies.

\end{abstract}

\begin{figure}[h]
  \centering
  \includegraphics[width=\linewidth]{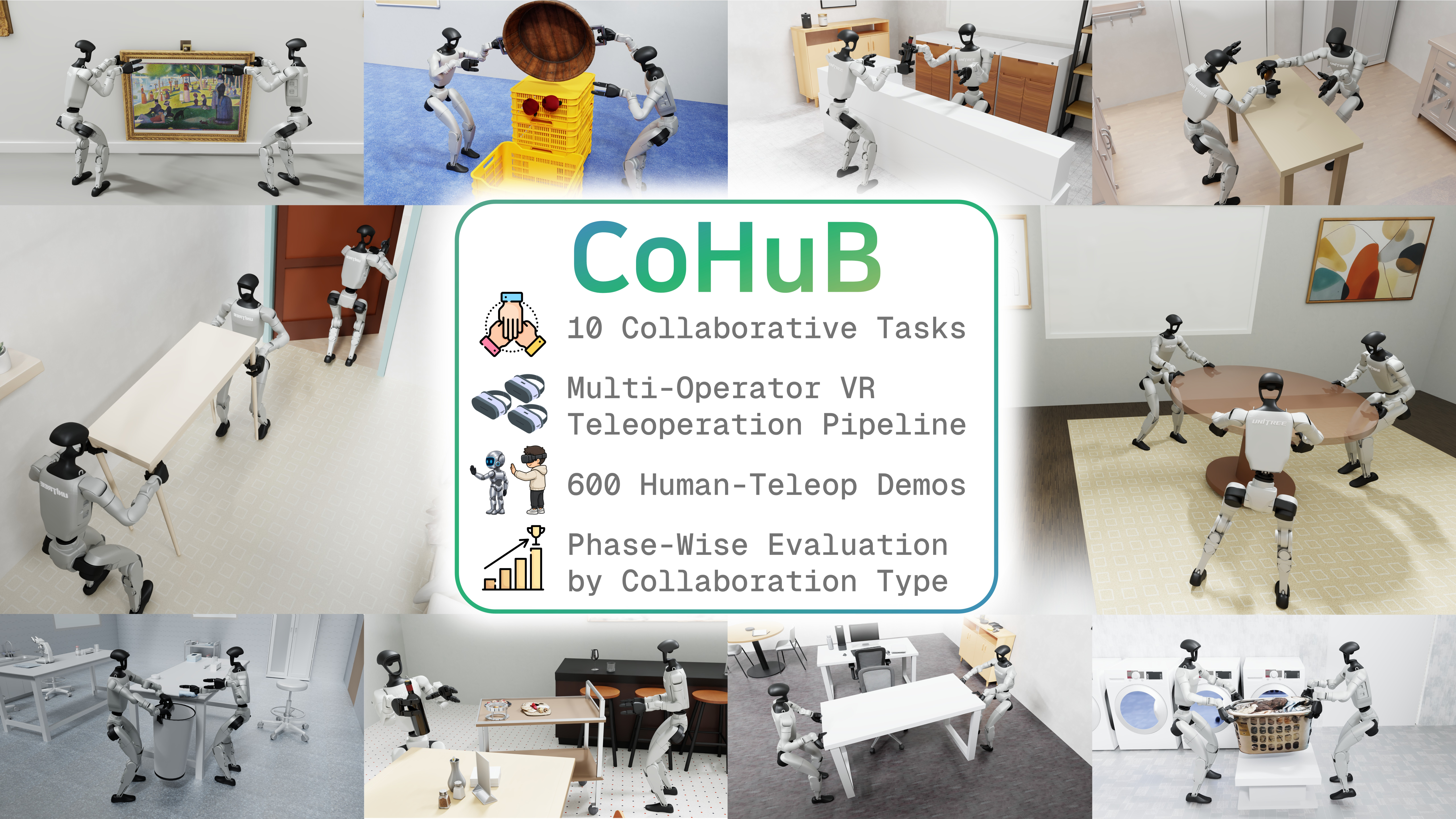}
    \caption{\textbf{Overview of \bench.}
    \bench is a simulation benchmark for multi-humanoid collaboration, featuring 10 collaborative tasks with two or three humanoids, a multi-operator VR teleoperation pipeline for synchronized data collection, and phase-wise evaluation of policy performance.}
  \label{fig:teaser}
\end{figure}

\section{Introduction}
Humanoid robots provide a suitable platform for operating in human environments, where spaces, tools, and tasks are designed around human morphology and capabilities.
Motivated by this potential, humanoid robot learning has advanced rapidly, with progress in whole-body control~\citep{luo2026sonic,xue2025leverb,ze2026twist2,ben2025homie}, foundation models~\citep{bjorck2025gr00t, wei2026psi0, zheng2026motionwam}, and simulation benchmarks~\citep{sferrazza2024humanoidbench,wei2026simple,wang2026humanoidarena}. These advances have made humanoids increasingly capable, yet the field remains largely centered on single-robot settings, even though many physical tasks in human environments inherently require collaboration.
For example, a long or heavy object may need to be held from multiple sides for stability, and someone with their hands full may need another person to open a door.
Such collaboration is harder than single-robot manipulation: shared objects can transmit forces between robots, potentially disturbing their balance. Successful execution may also depend on roles and coordinated timing between agents.

Developing policies in such collaborative settings requires demonstrations to learn from and a shared environment for training and evaluation, both of which are especially costly and difficult to reproduce in the real world with multiple humanoids.
Simulation offers a scalable and reproducible alternative. However, existing simulation benchmarks for humanoids focus on single-robot settings and do not evaluate collaboration among multiple humanoids~\citep{sferrazza2024humanoidbench,lin2026humanoidmimicgen,wei2026simple,wang2026humanoidarena}, while existing multi-agent benchmarks with visual observations have primarily focused on non-humanoid, table-top manipulation~\citep{qin2025robofactory}.

To this end, we present \bench (\textbf{Co}llaborative Multi-\textbf{Hu}manoid \textbf{B}enchmark), a simulation benchmark for multi-humanoid collaboration. Built on Isaac Lab~\citep{mittal2025isaaclab}, \bench features Unitree G1 humanoids equipped with Dex3-1 three-finger hands.
\bench includes 10 collaborative tasks: 8 tasks with two humanoids and 2 tasks with three humanoids, spanning diverse patterns of multi-humanoid collaboration.
Each task is decomposed into sequential phases to identify collaboration-critical phases and evaluate policy performance at each phase.
\bench also provides human demonstrations collected through synchronized multi-operator VR teleoperation, where each operator controls one humanoid in a shared simulation from its egocentric view.
We evaluate representative standard and multi-agent imitation learning (IL) policies, alongside Vision-Language-Action (VLA) models and a World Action Model (WAM). We further examine how their performance changes across task phases and how multi-agent design choices affect the performance. We also test GPT-6 Astra~\citep{openai2026astra} as an agentic robot policy in an exploratory evaluation.

The contributions of our work are summarized as follows:
\begin{itemize}[leftmargin=10pt]
    \item We introduce \bench, a simulation benchmark for multi-humanoid collaboration under egocentric visual observations, featuring a diverse task suite and a phase-wise evaluation protocol for analyzing collaboration requirements across task phases.
    \item We develop a synchronized multi-operator VR teleoperation pipeline that connects multiple operators to a shared simulation and release the collaborative demonstrations collected through it.
    \item We evaluate representative visuomotor policies, analyze their performance across task phases, and investigate the effects of different multi-agent policy configurations.
\end{itemize}

\section{Related Work}
\setlength{\textfloatsep}{10pt plus 0pt minus 2pt}
\begin{table}[t]
  \begingroup
  \renewcommand{\yes}{\textcolor{green!60!black}{\ding{51}}}
  \renewcommand{\no}{\textcolor{red!75!black}{\ding{55}}}
  \caption{
    \textbf{Comparison of representative simulation-based robot policy learning benchmarks.}
    \emph{Whole-body} indicates whether tasks require coordinated locomotion and manipulation using the whole body.
    \emph{Collaboration} indicates whether multiple embodied agents must collaborate to accomplish a shared task.
    \emph{Human teleop demos} indicates whether demonstrations collected through human teleoperation are provided.
    \yes yes, \no no.
  }
  \label{tab:related}

  \centering
  \footnotesize
  \setlength{\tabcolsep}{2.0pt}
  \renewcommand{\arraystretch}{1.10}

  \begin{tabular}{@{}llcccc@{}}
    \toprule
    \bfseries Benchmark
      & \bfseries Embodiment
      & \bfseries \# Agents
      & \makecell{\bfseries Whole-\\\bfseries body}
      & \bfseries Collaboration
      & \makecell{\bfseries Human Teleop\\\bfseries Demos} \\
    \midrule

    RLBench~\citep{james2020rlbench}
      & Manipulator & 1 & \no & \no & \no \\

    LIBERO~\citep{liu2023libero}
      & Manipulator & 1 & \no & \no & \yes \\

    RoboTwin 2.0~\citep{chen2026robotwin}
      & Manipulator & 1 & \no & \no & \no \\

    BEHAVIOR-1K~\citep{li2022behavior}
      & Mobile manip. & 1 & \no & \no & \yes \\

    HumanoidBench~\citep{sferrazza2024humanoidbench}
      & Humanoid & 1 & \yes & \no & \no \\

    SIMPLE~\citep{wei2026simple}
      & Humanoid & 1 & \yes & \no & \yes \\

    HumanoidArena~\citep{wang2026humanoidarena}
      & Humanoid & 1 & \yes & \no & \yes \\

    RoboFactory~\citep{qin2025robofactory}
      & Manipulator & 1--4 & \no & \yes & \no \\

    \midrule

    \textbf{\bench (ours)}
      & Humanoid & 2--3 & \yes & \yes & \yes \\

    \bottomrule
  \end{tabular}

  \endgroup
\end{table}

\noindent\textbf{Simulation Benchmarks for Robot Policy Learning.}
Simulation-based manipulation benchmarks have accelerated robot learning
by providing scalable and reproducible environments for policy training
and evaluation~\citep{zhu2020robosuite,yu2020meta,james2020rlbench,lee2021ikea,mees2022calvin,liu2023libero}.
Subsequent benchmarks have expanded the range of embodiments and task settings they cover.
PerAct$^2$~\citep{grotz2024peract2} and RoboTwin 2.0~\citep{chen2026robotwin} target bimanual manipulation, while BiGym~\citep{chernyadev2024bigym} extends benchmarking to mobile manipulation. BEHAVIOR-1K~\citep{li2022behavior} and RoboCasa~\citep{nasiriany2024robocasa} broaden it to diverse everyday activities in large-scale household environments.
A growing line of benchmarks further considers humanoid embodiments and whole-body interaction, while remaining primarily single-agent
~\citep{sferrazza2024humanoidbench,lin2026humanoidmimicgen,wei2026simple,wang2026humanoidarena}.
RoboFactory~\citep{qin2025robofactory} extends simulation-based
benchmarking to collaborative multi-agent settings with visual observations, but focuses on multi-arm embodiments and a narrow range of collaborative interactions.
Table~\ref{tab:related} summarizes these differences across representative benchmarks.
Thus, humanoid and multi-agent benchmarking have largely evolved separately, with limited support for evaluating diverse forms of collaborative humanoid interaction.

\noindent\textbf{Multi-Robot Learning and Collaboration.}
Learning-based approaches have increasingly enabled multiple robots to collaborate toward shared goals.
GauDP~\citep{wang2025gaudp} studies centralized diffusion policies for multi-agent collaboration, while LatentToM~\citep{he2025latent} and MIMIC-D~\citep{dong2026mimic} focus on decentralized execution.
CHORUS~\citep{doshi2026chorus} further extends this direction to pretrained Vision-Language-Action policies.
These methods, however, are primarily demonstrated with manipulators rather than humanoid robots.
Prior work has also explored collaboration among multiple humanoids.
CooHOI~\citep{gao2024coohoi} and TeamHOI~\citep{lionar2026teamhoi} use Multi-Agent Reinforcement Learning (MARL) for collaborative object transportation, but focus on physics-based humanoid characters with state-based observations.
Meanwhile, Harmanoid~\citep{liu2025takes} and Rhythm~\citep{chen2026rhythm} consider systems with two humanoid robots, but focus on tracking interactive whole-body paired motions, such as dancing or hugging, rather than learning task-oriented collaborative loco-manipulation.
Together, these works suggest that task-oriented physical collaboration among multiple humanoid robots remains underexplored.

\noindent\textbf{Robot Foundation Models and World Action Models.}
Recent robot foundation models have shown strong generalization across diverse manipulation tasks, objects, and environments~\citep{kim2024openvla,black2024pi_0,intelligence2025pi_}.
Humanoid-oriented foundation models have further extended this direction to humanoid loco-manipulation~\citep{bjorck2025gr00t,wei2026psi0}.
More recently, World Action Models (WAMs) have emerged as robot policies that leverage learned world dynamics for action generation~\citep{ye2026world,kim2026cosmos,yuan2026fast,zheng2026motionwam}.
Despite these advances, both robot foundation models and WAMs have been developed and evaluated primarily in single-agent settings.
It remains unclear whether humanoids controlled by these models can effectively collaborate with other humanoids and account for their partners' behavior.

\section{\bench: A Simulation Benchmark for Multi-Humanoid Collaboration}
\label{sec:benchmark}

In this section, we present \bench, which aims to support demonstration collection, policy learning, and evaluation for multi-humanoid collaboration.
We describe its simulated environment (\S\ref{sec:env}), task design principles and taxonomy (\S\ref{sec:principles}), task suite (\S\ref{sec:taxonomy}), multi-operator VR teleoperation system and dataset (\S\ref{sec:teleop}), and evaluation protocol (\S\ref{sec:evaluation}).

\subsection{Benchmark Environment}
\label{sec:env}

\bench is built on Isaac Lab~\citep{mittal2025isaaclab}, which runs on NVIDIA Isaac Sim.
\bench provides a unified environment across tasks, sharing the same humanoid embodiment, action and observation spaces, and data-recording pipeline.
Tasks differ only in their scene assets and success criteria.
Physics is simulated with PhysX at 200\,Hz, agents are controlled at 50\,Hz, and camera observations are rendered using Isaac Sim's RTX ray-tracing renderer.

\noindent\textbf{Scene assets and extensibility.}
Rather than authoring each environment from scratch, we instantiate task environments from publicly available simulation-ready scenes. These include scenes from SceneSmith~\citep{pfaff2026scenesmith}, which generates simulation-ready indoor scenes from natural-language prompts. We keep scene assets decoupled from the robots and task logic, enabling new environments with lightweight adaptation, such as repositioning furniture or task-relevant objects.

\noindent\textbf{Embodiment and action space.}
\bench uses Unitree G1 humanoids equipped with Unitree Dex3-1 three-finger hands.
We select G1 as a widely adopted research platform with established whole-body control models, and Dex3-1 as a balance between grasp versatility and control complexity.
Following the decoupled whole-body control interface used for G1 in GR00T~\citep{bjorck2025gr00t,nvidia2026grootwbc}, the upper body is commanded by joint targets, while a pretrained locomotion policy~\citep{zhao2026agile} executes lower-body motion from a base command. Demonstrations and all baselines use the same per-agent interface across tasks.

\noindent\textbf{Observation space.}
Each robot observes only its own egocentric RGB image and proprioception by default, together with a language instruction for VLA/WAM baselines. We choose this observation space to reflect onboard sensing available in real-world deployment and to evaluate collaboration from each humanoid's egocentric view. The demonstrations also retain synchronized auxiliary observations and simulator signals, including third-person views, depth, object poses, object grasp status, and wrist force--torque measurements, which are excluded from default baseline training. Appendix~\ref{app:plan} provides the full robot, controller, camera, and action specifications.

\subsection{Task Design Principles and Taxonomy}
\label{sec:principles}

A benchmark for multi-humanoid collaboration should capture diverse forms of interaction.
We therefore design our task suite along two dimensions:
\emph{base movement} and \emph{physical coupling}, motivated by prior taxonomies of interactive and collaborative behaviors~\citep{jarrasse2012framework,krebs2022bimanual}.
These axes describe the movement and physical interaction required by a task.

\noindent\textbf{Base movement.}
Base movement describes how much the humanoids need to move from their
initial positions during task execution.
We classify a task as low base movement when it can be completed near the humanoids' starting positions, with only small steps or local position adjustments.
In contrast, high base movement tasks require one or more humanoids to walk to another part of the scene or move between spatially separated work areas as part of completing the task.
For example, an interaction that can be completed while two humanoids stand near the same table is considered low base movement, whereas a task that requires carrying an object to another area of the room
is considered high base movement.

\noindent\textbf{Physical coupling.}
Physical coupling describes how much the humanoids must physically interact with the same object during a collaborative task.
We classify a task as low coupling when simultaneous interaction is absent or only brief, and the humanoids perform most of their actions independently or in sequence.
For example, in an object handover, both humanoids may briefly hold the object at the same time, but only during the transfer.
In contrast, high coupling tasks require multiple humanoids to maintain simultaneous physical interaction with the same object for a sustained portion of the task.
Since the humanoids manipulate the shared object at the same time, their motions remain physically linked through the object.
Examples include jointly lifting, carrying, or aligning a large object, where sustained co-manipulation is required for task completion.

\begin{figure}[t]
  \centering
  \includegraphics[width=\textwidth]{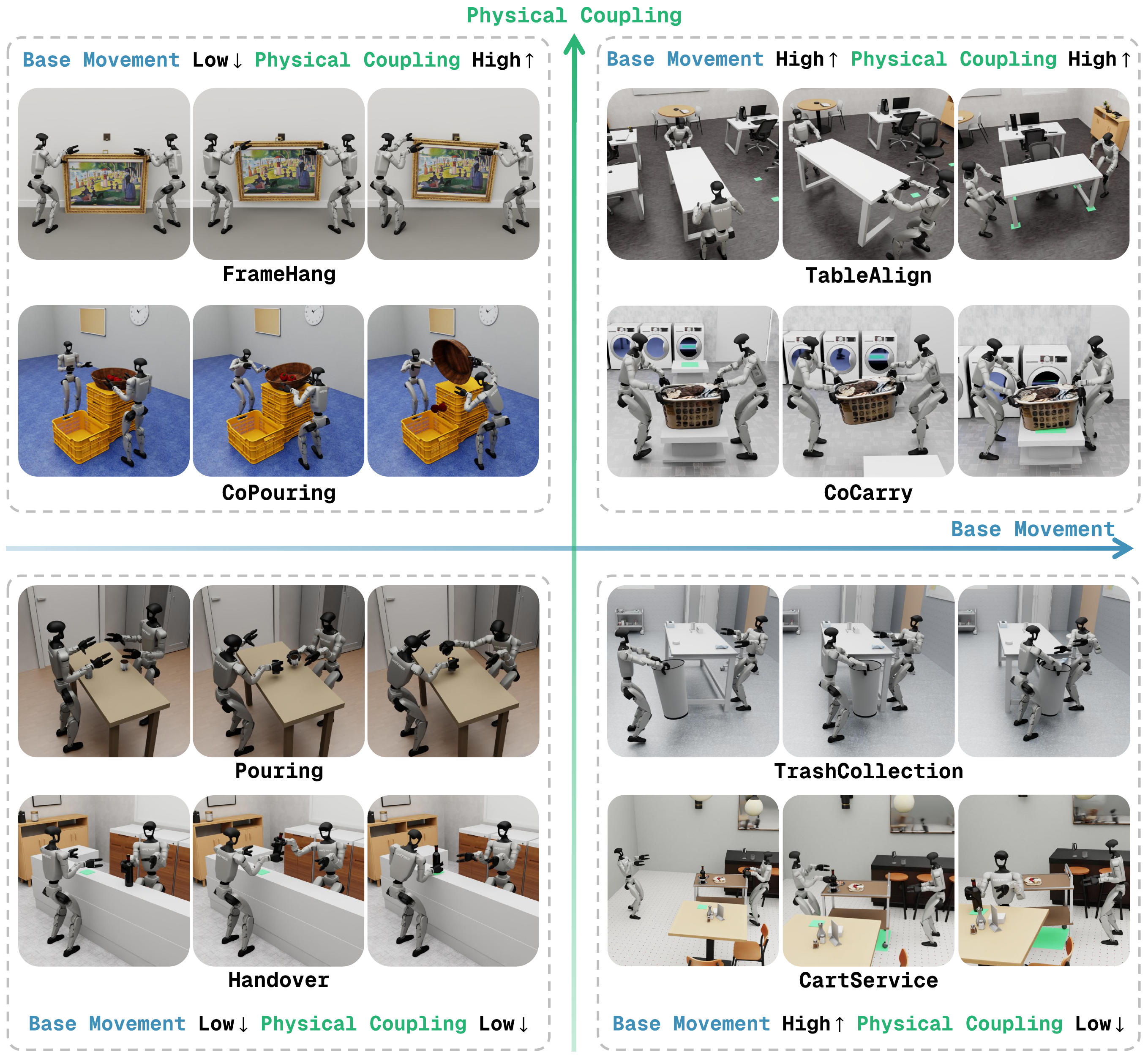}
  \par\vspace{-7pt}
  \caption{\textbf{The \bench task taxonomy.} The 8 tasks with two humanoids are organized by base movement (horizontal axis) and physical coupling (vertical axis), with two tasks in each quadrant. Each task is illustrated by three key frames ordered from left to right.} 
  \label{fig:tasks-overview}
\end{figure}

Together, these two dimensions define four task categories, with two tasks designed for each category, as illustrated in Figure~\ref{fig:tasks-overview}.
These categories do not imply that one is more difficult than another.

\subsection{Task Suite}
\label{sec:taxonomy}

\bench includes 8 tasks with two humanoids across the four categories and 2 additional tasks with three humanoids.
Figures~\ref{fig:gallery-two-low}--\ref{fig:gallery-three} in Appendix~\ref{app:tasks} show key frames of all 10 tasks.

\noindent\textbf{Low base movement \& Low physical coupling.}
In \emph{Handover},
one humanoid transfers a bottle to its partner,
which then places it at the target location.
In \emph{Pouring},
one humanoid pours a marble from a small cup into a glass held by its partner.

\noindent\textbf{Low base movement \& High physical coupling.}
In \emph{FrameHang},
two humanoids jointly lift and align a large picture frame
and hang it on a wall-mounted peg.
In \emph{CoPouring},
they lift a tub of apples together and tilt it
to pour the apples into an empty crate.

\noindent\textbf{High base movement \& Low physical coupling.}
In \emph{TrashCollection},
one humanoid carries a bin to the end of a table,
where its partner sweeps a bottle into it.
In \emph{CartService},
one humanoid pushes a trolley to a table,
while its partner retrieves a wine bottle from the trolley
and places it on the table.

\noindent\textbf{High base movement \& High physical coupling.}
In \emph{CoCarry},
two humanoids jointly carry a laundry basket
to a target location while keeping it balanced.
In \emph{TableAlign},
they carry a table across the office,
rotate it during transport,
and place it aligned with a chair.

\noindent\textbf{Tasks with three humanoids.}
In \emph{MoveHouse},
one humanoid opens a door while the other two carry a table
through the doorway and place it at the target location.
In \emph{BigTable},
all three humanoids jointly lift a large round table,
carry it to the target location,
and set it down together.

\begin{figure}[t]
  \centering
  \includegraphics[width=\linewidth]{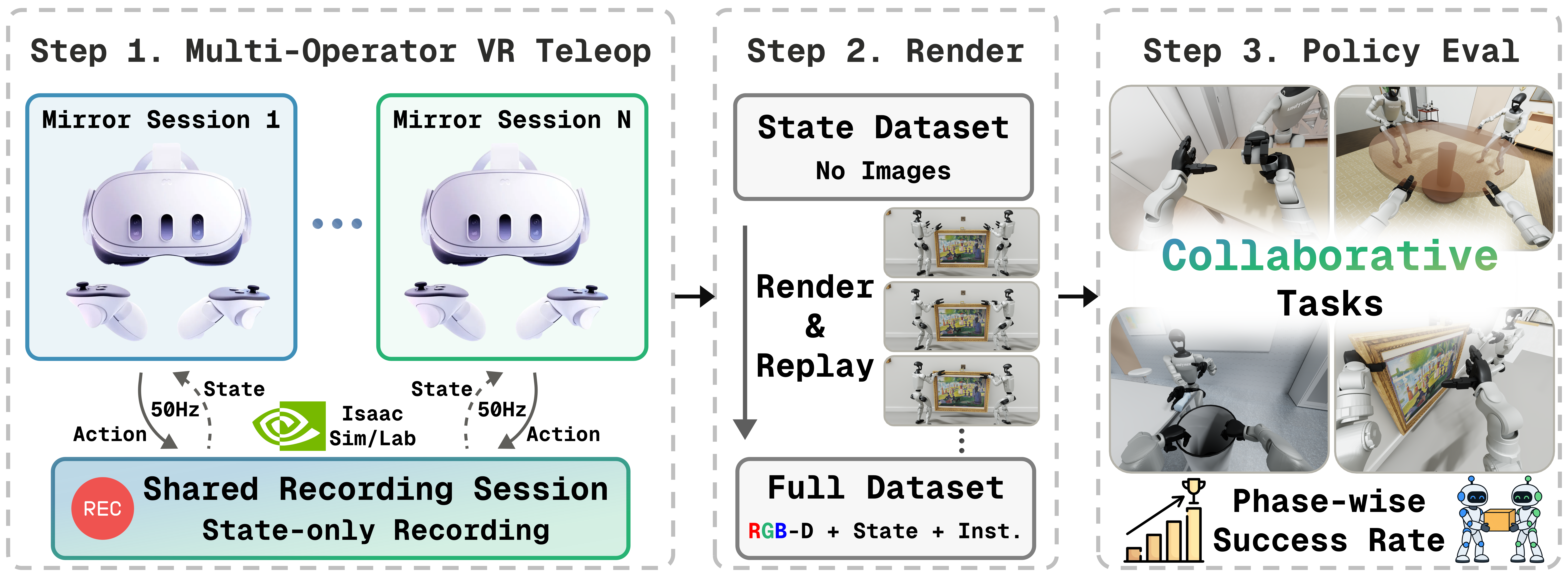}
  \par\vspace{-7pt}
  \caption{\textbf{\bench pipeline.} The multi-operator teleoperation system records synchronized state trajectories, which are replayed offline to render RGB-D observations and construct the full demonstration dataset for policy training and evaluation.}
  \label{fig:pipeline}
\end{figure}

\subsection{Multi-Operator VR Teleoperation and Dataset}
\label{sec:teleop}

Collecting collaborative demonstrations with multiple operators requires synchronized control and low-latency feedback so that operators can react to one another in real time.
However, recording multiple camera streams online can increase runtime overhead.
To address these challenges, we develop a multi-operator VR teleoperation system that supports synchronized control while reducing online recording overhead.

\noindent\textbf{Egocentric VR teleoperation.}
We extend Isaac Teleop~\citep{isaacteleop2026} from its single-operator setting to support multiple operators, each controlling one humanoid from its egocentric view within a shared, synchronized simulation.
The operators' commands are applied together at each control step, enabling coordinated interaction with real-time visual feedback.
During data collection, the VR interface displays the boundary of the humanoid's onboard camera field of view to the operator, encouraging coordination based on visual information available to the policy during execution.

\noindent\textbf{Multi-operator architecture and dataset construction.}
Our multi-operator teleoperation pipeline consists of two stages.
During Step~1 in Figure~\ref{fig:pipeline}, only states and actions are recorded online, avoiding the overhead of recording multiple camera streams.
Each operator uses a separate local \emph{mirror} session, since each headset requires its own Isaac Sim instance.
The operator controls one humanoid through this session and receives its egocentric VR view.
However, running physics independently in each mirror session could lead to inconsistent states across humanoids.
We therefore run a single \emph{shared recording} session to maintain synchronized physics and states across all humanoids.
Each mirror synchronizes with the latest states from the shared session, while operator commands are sent back and applied together.
During Step~2, the trajectories are replayed to render image observations, producing synchronized visual observations, states, and actions across humanoids.
Across the 10 tasks, the resulting dataset contains 600 synchronized demonstrations, with 50 training and 10 validation demonstrations per task. In total, the dataset comprises 1,320 per-humanoid trajectories. 
Details on the train--validation split and validation usage are provided in Appendix~\ref{app:data-split}.

\subsection{Evaluation Protocol}
\label{sec:evaluation}

To support standardized comparison and failure analysis, we use two evaluation protocols: full-task rollouts for the main benchmark results and a phase-level evaluation for analyzing collaboration. Both protocols share the same success criteria.

\noindent\textbf{Success criteria.}
Each task is divided into three sequential phases ($p_1 \rightarrow p_2 \rightarrow p_3$), each with task-specific completion criteria based on robot and object states.
For example, a phase may require an object to be lifted above a certain height, held by specific humanoids, or placed within a target region.
A phase is considered successful only after all earlier phases have been completed, and task success is defined by completion of $p_3$.
Detailed criteria for each task are listed in Table~\ref{tab:app-task-phases}.

\noindent\textbf{Main benchmark evaluation.} 
For the main results in Table~\ref{tab:main-compact}, each policy is trained with a single random seed, and the resulting checkpoint is evaluated over 100 rollouts per task.
Each rollout starts from the beginning of the task in a fixed scene, with the humanoids initialized at fixed poses, while task object and target poses are sampled uniformly from task-specific ranges (Table~\ref{tab:app-scene}).
We record both task success and the completion of each phase.

\noindent\textbf{Phase-level evaluation.}
We categorize each phase as either
\textcolor{noncollabphase}{\textbf{non-collaborative}} or
\textcolor{collabphase}{\textbf{collaborative}},
depending on whether its completion requires interaction between humanoids.
For example, in FrameHang, grasping each side of the frame is non-collaborative,
whereas jointly lifting the frame and aligning its loop with the wall-mounted peg is collaborative.
To compare these phases while reducing the influence of failures accumulated earlier in the task, we conduct a separate phase-level evaluation.
Unlike the main benchmark evaluation, this evaluation does not start from the beginning of the tasks.
Each rollout starts from one of the 50 training demonstrations where the evaluated phase begins, yielding 50 rollouts per phase.
The resulting average success rates of non-collaborative and collaborative phases are reported in Figure~\ref{fig:collab-cond-gap}.

\section{Experiments}
\label{sec:results}

We now evaluate representative visuomotor policies on the \bench task suite to assess their performance and identify challenges in multi-humanoid collaboration, along with an exploratory evaluation of an agentic robot policy.
We first describe the baselines and experimental settings (\S\ref{sec:exp-setup}), then present benchmark results with qualitative analysis (\S\ref{sec:overall-results}).
Next, we compare the baselines' performance across phases with different collaboration requirements (\S\ref{sec:collab-results}).
Finally, we analyze how multi-agent design choices affect performance (\S\ref{sec:policy-design}).

\newcolumntype{R}[2]{%
  >{\adjustbox{angle=#1,lap=\width-(#2)}\bgroup}%
  l%
  <{\egroup}%
}
\newcommand*{\rot}{\multicolumn{1}{R{30}{1em}}}
\newcommand{\best}[1]{\textbf{#1}}

\definecolor{catrow}{RGB}{232,238,245}
\definecolor{catrowmt}{RGB}{230,245,230}
\newcommand{\famrowstrip}[3]{%
  \multicolumn{#2}{@{}l}{%
    \setlength{\fboxsep}{0pt}%
    \makebox[0pt][l]{%
      \colorbox{#1}{%
        \makebox[\linewidth][l]{%
          \rule[-0.6ex]{0pt}{2.5ex}\hspace{3pt}\emph{#3}%
        }%
      }%
    }%
  }\\
}
\newcommand{\famrow}[2]{\famrowstrip{catrow}{#1}{#2}}
\newcommand{\famrowmt}[2]{\famrowstrip{catrowmt}{#1}{#2}}

\input{tables/main-results-table}

\subsection{Experimental Setup and Baselines}
\label{sec:exp-setup}

\noindent\textbf{Main baselines.}
We benchmark 8 visuomotor policies grouped into 4 representative categories.
\textbf{Standard IL} includes Action Chunking with Transformer (ACT)~\citep{zhao2023learning} and Diffusion Policy (DP)~\citep{chi2025diffusion}, and
\textbf{Multi-agent IL} includes LatentToM~\citep{he2025latent} and GauDP~\citep{wang2025gaudp}.
We use $\pi_{0.5}$~\citep{intelligence2025pi_}, GR00T N1.7~\citep{bjorck2025gr00t}, and $\Psi_0$~\citep{wei2026psi0} for the \textbf{VLAs}, and Fast-WAM~\citep{yuan2026fast} as the \textbf{WAM} baseline.

\noindent\textbf{Training and evaluation settings.}
We use different training and execution configurations across policy families:
\begin{itemize}[leftmargin=10pt, itemsep=1pt, topsep=2pt]
    \item \textbf{Standard IL.}
    ACT and DP are trained separately for each task and humanoid.
    At execution, each humanoid's policy predicts its action from egocentric RGB and proprioception.

    \item \textbf{Multi-agent IL.}
    LatentToM uses centralized training and decentralized execution, with each policy observing its own egocentric observation and a shared third-person view.
    GauDP uses centralized training and execution, taking all agents' egocentric observations as input and jointly predicting their actions.

    \item \textbf{VLA and WAM.}
    Each model is fine-tuned from a pretrained checkpoint in a multi-task setting, separately over the two- and three-humanoid task suites.
    All humanoids share the same policy parameters, and the policy is queried separately for each humanoid using its egocentric RGB image, proprioception, and agent-specific instruction.
\end{itemize}


\noindent\textbf{Agentic robot policy.}
GPT-6 Astra~\citep{openai2026astra} controls both humanoids using egocentric RGB images and proprioception, with access to depth queries at selected pixels.
It outputs base, hand, and wrist commands.
We allow a larger episode time budget to support reasoning and replanning.


Further details for all methods are provided in Appendix~\ref{app:eval}.

\subsection{Overall Benchmark Results}
\label{sec:overall-results}

\noindent\textbf{Quantitative results.}
Table~\ref{tab:main-compact} summarizes overall success rates across tasks with two and three humanoids.
Among the IL-based baselines, ACT achieves the strongest performance, substantially outperforming DP across most tasks.
Despite explicitly modeling multi-agent interactions, LatentToM and GauDP achieve lower task success than ACT.
Among the pretrained models, Fast-WAM achieves the highest average success rates.
In contrast, the VLAs achieve relatively low success rates, suggesting that adapting pretrained VLAs to multi-humanoid collaboration remains challenging under our evaluated settings.
Phase-wise success rates are provided in Appendix~\ref{app:phasewise}.

\noindent\textbf{Qualitative analysis.}
We further analyze policy rollouts to characterize recurring behaviors and failure modes beyond the predefined success metrics, revealing distinct failure patterns across tasks (Appendix~\ref{app:failures}). For example, Handover yields the same 2\% success rate for $\pi_{0.5}$ and Fast-WAM despite different failure patterns.
$\pi_{0.5}$ often collides with its partner's arm during the transfer ($p_2$), whereas Fast-WAM more often completes the transfer but fails to release the bottle ($p_3$). CoCarry and TableAlign also exhibit a different pattern. VLAs often complete the joint lift ($p_1$) but drop the object during transport ($p_2$), whereas Fast-WAM more often maintains its grasp but the two humanoids move inconsistently while carrying the object ($p_2$). These cases show that \bench requires both capable single-agent behavior and reliable multi-agent collaboration.

\noindent\textbf{Agentic model evaluation.}
GPT-6 Astra's performance varies substantially across two-humanoid tasks, with 70\% on CoPouring and 10\% on Pouring, while achieving no successes on the remaining tasks.
Although its separate observation, control, and time-budget settings limit direct comparison with the trained baselines, this contrasting profile suggests different strengths and limitations from learned visuomotor policies.
Details are provided in Appendix~\ref{app:astra-evaluation}.

\input{figures/fig-collab-cond-gap}

\subsection{How Does Performance Differ Between Non-Collaborative and Collaborative Phases?}
\label{sec:collab-results}

To better understand the benchmark results, we compare policy performance across phases with different collaboration requirements using the phase-level evaluation described in Section~\ref{sec:evaluation}.
Figure~\ref{fig:collab-cond-gap} shows consistently lower success rates
in \textcolor{collabphase}{\textbf{collaborative}} phases than in
\textcolor{noncollabphase}{\textbf{non-collaborative}} phases across all policy families,
suggesting that collaboration is associated with lower performance within our task suite.
Specifically, LatentToM and GauDP achieve lower collaborative phase success rates (56.1\% and 60.6\%) than ACT (72.1\%).
Among the VLA and WAM baselines, Fast-WAM achieves the highest collaborative phase success rate (62.7\%).
Fast-WAM and $\pi_{0.5}$ perform comparably on non-collaborative phases (76.8\% vs.\ 76.6\%), yet Fast-WAM outperforms $\pi_{0.5}$ by 10.3 percentage points (62.7\% vs.\ 52.4\%) on collaborative phases.
One possible explanation is that Fast-WAM's video co-training may help capture temporal interaction dynamics~\citep{yuan2026fast}.
We note that the observed differences may also reflect variations in phase difficulty.

\subsection{How Do Multi-Agent Design Choices Affect Performance?}
\label{sec:policy-design}
\input{tables/obs-act-table}

Given that explicit multi-agent methods do not necessarily outperform standard IL baselines, we compare different configurations of ACT, DP, and GR00T N1.7, the best-performing VLA on the two-humanoid task suite.
Following RoboFactory~\citep{qin2025robofactory}, we vary observation scope and whether policy parameters are shared across humanoids.

\noindent\textbf{Observation scope.}
\emph{Local Observation} uses the controlled humanoid's egocentric RGB image and proprioception.
\emph{Global Observation} uses the egocentric RGB images and proprioception of all humanoids, rather than an external-camera view as in RoboFactory's Global View.

\noindent\textbf{Policy.}
\emph{Separate Policies} use distinct policy parameters for each humanoid, whereas a \emph{Shared Policy} uses the same parameters for all humanoids.

Combining the two factors yields 4 configurations for each model (Table~\ref{tab:view-policy}).
For ACT and DP, no agent identifier is provided in the \emph{Local Observation} + \emph{Shared Policy} configuration.
For GR00T N1.7, agent-specific instructions are provided in all configurations except \emph{Global Observation} + \emph{Shared Policy}, where a joint instruction is used.
We compare these configurations within each model under matched data, training budgets, and evaluation conditions.

\noindent\textbf{Results.}
For both ACT and DP, \emph{Local Observation} + \emph{Shared Policy} yields the lowest average success (23.9\% and 9.5\%) among the four configurations.
This may reflect the difficulty of distinguishing between humanoids using only local observations without an explicit identifier, as discussed in RoboFactory~\citep{qin2025robofactory}.
For DP, \emph{Global Observation} improves performance under both policy settings (19.1\% vs.\ 10.4\% with \emph{Separate Policies}, 16.0\% vs.\ 9.5\% with a \emph{Shared Policy}).
For GR00T N1.7, \emph{Separate Policies} achieve higher average success than a \emph{Shared Policy} under both \emph{Local Observation} (14.5\% vs.\ 9.4\%) and \emph{Global Observation} (12.3\% vs.\ 9.0\%).
\emph{Local Observation} also achieves higher success than \emph{Global Observation} under both \emph{Separate Policies} (14.5\% vs.\ 12.3\%) and a \emph{Shared Policy} (9.4\% vs.\ 9.0\%).
We hypothesize that these results may reflect distribution shift relative to GR00T N1.7's pretraining setting. Specifically, a \emph{Shared Policy} must handle different humanoid roles, while \emph{Global Observation} provides multiple egocentric views, both of which differ from those seen during pretraining. 

In summary, our findings show that the effects of observation scope and policy sharing vary across models, providing useful insights for future research in multi-humanoid collaboration.

\section{Conclusion}

In this work, we introduced \bench, a simulation benchmark for multi-humanoid collaboration that covers diverse collaboration patterns, supports synchronized data collection through multi-operator VR teleoperation, and enables phase-wise analysis of collaborative performance.
Our experiments show that current policies remain limited in multi-humanoid settings, with lower performance observed during collaborative phases.
These results suggest that successful multi-humanoid collaboration requires policies that can better respond to other humanoids during interaction.
We believe \bench provides a reproducible platform for developing and evaluating policies across diverse patterns of multi-humanoid collaboration.
Future work could expand the task suite to capture aspects of collaboration beyond our current taxonomy, such as explicit inter-agent communication and long-horizon tasks requiring high-level planning.

\subsection*{AI use statement}

In this work, we used generative AI tools for proposing or refining hypotheses, designing or providing feedback on research methodology and experiments, and implementing methods.
Additionally, we used generative AI tools for creating or editing software code, suggesting experimental parameters, creating or modifying scientific figures or images, drafting parts of a research paper, editing a research paper to improve readability, summarizing or analysing existing literature, identifying relevant literature, sourcing/searching for information, and proposing a title or keywords for a research paper. We have reviewed all AI-assisted work. We reviewed and tested AI-assisted code for correctness, and reviewed AI-assisted text for accuracy and consistency.
We take responsibility for the final content of this work, including text, claims or artifacts produced with the aid of generative AI.

\bibliographystyle{iclr2027_conference}
\bibliography{references}

\appendix

\newpage

\newcommand{\apptocsection}[2]{%
  \par\addvspace{1.0em}%
  \noindent
  {\large\textbf{\hyperref[#1]{\ref*{#1}.\ #2}}}
  \dotfill \raisebox{-0.1ex}{\pageref{#1}}\par
}

\newcommand{\apptocsubsection}[2]{%
  \noindent\hspace*{1.5em}%
  {\normalsize\hyperref[#1]{\ref*{#1}\ #2}}
  \dotfill \raisebox{-0.1ex}{\pageref{#1}}\par
}

{\Large\bfseries Appendix}\par
{\large\bfseries Table of Contents}

\vspace{1.2em}

\begingroup
\hypersetup{hidelinks}
\setlength{\parskip}{0.8em}

\apptocsection{app:environment}
{Simulation Environment and Robot Setup}

\apptocsubsection{app:control}
{Action Space and Control}

\apptocsubsection{app:observation}
{Observations and Cameras}

\apptocsection{app:tasks}
{Task and Dataset Specifications}

\apptocsubsection{app:task-scenes}
{Task Scenes and Randomization}

\apptocsubsection{app:task-phases}
{Phase Definitions and Success Criteria}

\apptocsubsection{app:instructions}
{Language Instructions}

\apptocsubsection{app:data-split}
{Demonstration Split and Usage}

\apptocsection{app:eval}
{Experimental Setup Details}

\apptocsubsection{app:baseline-inputs}
{Policy Inputs}

\apptocsubsection{app:baseline-training}
{Training Configurations}

\apptocsubsection{app:baseline-evaluation}
{Rollout Evaluation}

\apptocsubsection{app:astra-evaluation}
{Agentic Robot Policy Evaluation}

\apptocsection{app:phasewise}
{Phase-Wise Results}

\apptocsection{app:failures}
{Collaboration Failure Cases}

\endgroup
\clearpage

\section{Simulation Environment and Robot Setup}
\label{app:environment}
\label{app:plan}

Table~\ref{tab:app-env} summarizes the simulation settings and per-robot action and observation spaces.

\begin{table}[!ht]
  \centering
  \caption{\textbf{Shared simulation and robot settings.} Robot count depends on the task; action and state dimensions are per robot.}
  \label{tab:app-env}
  \footnotesize
  \renewcommand{\arraystretch}{1.15}
  \setlength{\tabcolsep}{6pt}
  \begin{tabularx}{\linewidth}{@{}l>{\raggedright\arraybackslash}X@{}}
    \toprule
    \textbf{Simulator}
    & Isaac Sim 6.0.1 with Isaac Lab \\
    \addlinespace[2pt]

    \textbf{Physics / env. control}
    & $200$\,Hz / $50$\,Hz \\
    \addlinespace[2pt]

    \textbf{Robots}
    & Two or three Unitree G1 humanoids, each with two Dex3-1 three-finger hands \\
    \addlinespace[2pt]

    \textbf{Joint DoF}
    & $43$: legs $12$, waist $3$, arms $14$, and hands $2\times7$ \\
    \addlinespace[2pt]

    \textbf{Action space}
    & $35$: upper-body joint targets $31$ and base command $4$ \\
    \addlinespace[2pt]

    \textbf{Proprioceptive state}
    & $43$ joint positions (rad) \\
    \addlinespace[2pt]

    \textbf{Egocentric camera}
    & RGB-D, $320\times240$ ($4{:}3$), $102^\circ$ horizontal field of view,
      $25^\circ$ downward tilt \\
    \bottomrule
  \end{tabularx}
\end{table}

\subsection{Action Space and Control}
\label{app:control}

The benchmark action space consists of $31$ absolute joint targets (rad) for
the arms, hands, and waist, and a base command $[v_x,v_y,\omega_z,h]$ specifying robot-frame planar velocity (m/s), yaw rate (rad/s), and pelvis height (m).
The $12$ leg joints remain part of the state but are generated by a pretrained AGILE locomotion controller~\citep{zhao2026agile} given the base command, rather than predicted directly by the task policy.
Environment actions are applied at $50$\,Hz, with four physics steps per control step. Policy inference can produce multiple actions as a chunk.

During teleoperation, each robot instead receives a $32$-D input: two wrist poses ($14$-D), hand joint targets ($14$-D), and the base command ($4$-D).
Pink~\citep{caron2026pink} converts the wrist poses into upper-body joint targets.
These recorded targets and base commands provide the $35$-D action labels for learning.
During policy execution, the policy directly predicts joint targets, so inverse kinematics is not required.
Model-specific input and output differences are described in Appendix~\ref{app:eval}.

\subsection{Observations and Cameras}
\label{app:observation}

The default policy observation consists of the controlled robot's own egocentric RGB image and $43$ joint positions, with a role-specific language
instruction for language-conditioned models.
The egocentric camera is positioned at the robot's head-camera location, with the tilt chosen to include the hands and nearby partners.
The dataset additionally records synchronized depth and third-person RGB images, robot and object poses, grasp states, and wrist wrenches.
These additional signals are excluded from the default policy inputs.

\section{Task and Dataset Specifications}
\label{app:tasks}

\bench includes 8 tasks with two humanoids and 2 tasks with three humanoids.
This section specifies their scenes, randomization, phase criteria, language instructions, and the training--validation split of the collected demonstrations.
Figures~\ref{fig:gallery-two-low} and~\ref{fig:gallery-two-high} show key frames of the tasks with two humanoids, split by base movement, and Figure~\ref{fig:gallery-three} those of the tasks with three humanoids.

\input{figures/task-gallery}

\subsection{Task Scenes and Randomization}
\label{app:task-scenes}

Each task uses a fixed scene adapted from SceneSmith~\citep{pfaff2026scenesmith}.\footnote{\url{https://huggingface.co/datasets/nepfaff/scenesmith-example-scenes}.}
We retain each scene's semantic setting while adjusting furniture placement and adding or adapting task objects.
Table~\ref{tab:app-scene} summarizes the scene types, source IDs, task objects, and randomization of initial object poses and task targets during evaluation.
In the source IDs, \texttt{Room} and \texttt{House} denote single-room and multi-room scenes, respectively.
At evaluation reset, offsets are sampled uniformly from the listed ranges relative to default poses in task coordinates.

\FloatBarrier
\begingroup
\setlength{\LTcapwidth}{\linewidth}
\setlength{\tabcolsep}{3pt}
\renewcommand{\arraystretch}{1.12}
\begin{longtable}{@{}>{\raggedright\arraybackslash}p{0.165\linewidth}>{\raggedright\arraybackslash}p{0.2\linewidth}>{\raggedright\arraybackslash}p{0.17\linewidth}>{\raggedright\arraybackslash}p{\dimexpr0.465\linewidth-18pt\relax}@{}}
\caption{\textbf{Task scenes, objects, and randomization.} Source IDs refer to the source dataset before task-specific adaptation. Translation offsets are in meters.}\label{tab:app-scene}\\
\toprule
\textbf{Task} & \textbf{Scene (source ID)} & \textbf{Task objects} & \textbf{Eval randomization} \\
\midrule
\endfirsthead
\caption[]{(continued) \textbf{Task scenes, objects, and randomization.} Source IDs refer to the source dataset before task-specific adaptation. Translation offsets are in meters.}\\
\toprule
\textbf{Task} & \textbf{Scene (source ID)} & \textbf{Task objects} & \textbf{Eval randomization} \\
\midrule
\endhead
\bottomrule
\endfoot
Handover & Kitchen\newline\texttt{Room/scene\_084} & Wine bottle; target pad & Bottle and target pad independently: $x,y\in[-0.025,0.025]$. \\
\addlinespace[4pt]
Pouring & Kitchen\newline\texttt{Room/scene\_085} & Cup, glass, and marble & Cup and receiving glass independently: $x\in[-0.03,0.03]$, $y\in[-0.06,0.06]$, yaw $\pm30^\circ$; marble position within the cup: $x,y\in[-0.006,0.006]$. \\
\addlinespace[4pt]
FrameHang & Art gallery\newline\texttt{Room/scene\_136} & Framed painting; wall peg & Target peg and frame share $x\in[-0.05,0.05]$ and wall depth $y\in[-0.025,0.025]$; target peg $z\in[-0.04,0]$; extra frame offsets $x\in[-0.03,0.03]$, $y\in[-0.01,0.01]$. \\
\addlinespace[4pt]
CoPouring & Grocery store\newline\texttt{Room/scene\_102} & Tub of apples; receiving crate & Tub: $x\in[-0.02,0.02]$, $y\in[-0.01,0.01]$; receiving crate (target), independently: $x\in[-0.06,0.06]$, $y\in[-0.005,0.005]$; apples relative to tub: $x,y\in[-0.008,0.008]$. Supporting stacks follow their containers. \\
\addlinespace[4pt]
TrashCollection & Science laboratory\newline\texttt{Room/scene\_147} & Bin; plastic bottle on bench & $+y$ toward Robot A. Bin: $x\in[-0.15,0.15]$, $y\in[-0.15,-0.04]$. Bottle: $x\in[-0.10,0.10]$, $y\in[-0.04,0.12]$. \\
\addlinespace[4pt]
CartService & Restaurant\newline\texttt{Room/scene\_110} & Service trolley; wine bottle & $+y$ toward Robot A. Trolley and its contents: $x\in[-0.10,0.10]$, $y\in[-0.30,0.08]$. Goal: $y\in[-0.16,-0.12]$. \\
\addlinespace[4pt]
CoCarry & Laundromat\newline\texttt{Room/scene\_143} & Laundry basket; goal shelf & Basket: $x,y\in[-0.05,0.05]$. \\
\addlinespace[4pt]
TableAlign & Open-plan office\newline\texttt{Room/scene\_115} & Table; fixed goal chair & Table: $x,y\in[-0.05,0.05]$. \\
\addlinespace[4pt]
MoveHouse & Home: bedroom and living room\newline\texttt{House/scene\_098} & Console table; lever-handle door & Table: $x,y\in[-0.05,0.05]$; door initially closed and latched. \\
\addlinespace[4pt]
BigTable & Dining room\newline\texttt{Room/scene\_046} & Round pedestal table & Table: $x,y\in[-0.04,0.04]$, yaw $\pm4^\circ$. \\
\end{longtable}
\endgroup

\subsection{Phase Definitions and Success Criteria}
\label{app:task-phases}

Table~\ref{tab:app-task-phases} defines the three scored phases of each task.
Each phase requires its predecessors, so a correct final object pose alone is insufficient for task success ($p_3$).
A robot is considered to hold an object when at least one of its hands satisfies the contact-based grasp detector, unless a particular hand is specified.
Joint holding requires all designated robots to hold the object simultaneously.

For final placements in Handover, CartService, CoCarry, TableAlign, MoveHouse, and BigTable, the object must be within the task's position tolerances, tilted at most $15^\circ$ from upright, and not moving faster than $0.05$\,m/s linearly and $0.15$\,rad/s angularly. Release must last at least $0.2$\,s. FrameHang instead permits $35^\circ$ tilt, $0.20$\,m/s, and $0.80$\,rad/s, with the same release duration.

Every task with two humanoids except CartService has at least one collaborative phase.
In CartService, all phases are non-collaborative, but the task as a whole is collaborative because Robot B retrieves the bottle from the trolley that Robot A delivers.

\begingroup
\setlength{\LTcapwidth}{\linewidth}
\setlength{\tabcolsep}{5pt}
\renewcommand{\arraystretch}{1.12}
\begin{longtable}{@{}>{\raggedright\arraybackslash}p{0.165\linewidth}>{\raggedright\arraybackslash}p{\dimexpr0.835\linewidth-10pt\relax}@{}}
\caption{\textbf{Success criteria of each phase.} Phases follow $p_1 \rightarrow p_2 \rightarrow p_3$; $p_3$ denotes task success. Distances refer to the manipulated object unless otherwise stated. For the tasks with two humanoids, the \textcolor{noncollabphase}{\textbf{non-collaborative}} and \textcolor{collabphase}{\textbf{collaborative}} phases used in Figure~\ref{fig:collab-cond-gap} are shown in blue and green, respectively.}\label{tab:app-task-phases}\\
\toprule
\textbf{Task} & \textbf{Scored phases and conditions} \\
\midrule
\endfirsthead
\caption[]{(continued) \textbf{Success criteria of each phase.} Phases follow $p_1 \rightarrow p_2 \rightarrow p_3$. $p_3$ denotes task success. Distances refer to the manipulated object unless otherwise stated. For the tasks with two humanoids, the \textcolor{noncollabphase}{\textbf{non-collaborative}} and \textcolor{collabphase}{\textbf{collaborative}} phases used in Figure~\ref{fig:collab-cond-gap} are shown in blue and green, respectively.}\\
\toprule
\textbf{Task} & \textbf{Scored phases and conditions} \\
\midrule
\endhead
\bottomrule
\endfoot
Handover & \noncollabphase{\textbf{$p_1$ Grasp and lift:}} A's right hand holds the bottle at least 3\,cm above its counter resting height.
\collabphase{\textbf{$p_2$ Transfer:}} A's and B's right hands hold it together, followed by B's right hand holding with both of A's hands off, without an intervening all-hands release.
\noncollabphase{\textbf{$p_3$ Place:}} Released, stable, upright placement within $\pm6$\,cm of the target on each axis. \\
\addlinespace[5pt]
Pouring & \noncollabphase{\textbf{$p_1$ Lift both vessels:}} Cup and glass are simultaneously at least 3\,cm above the worktop.
\noncollabphase{\textbf{$p_2$ Tip:}} A holds the cup aloft and tilts it at least $60^\circ$.
\collabphase{\textbf{$p_3$ Pour:}} The marble remains inside the glass continuously for 0.5\,s, with the glass at least 3\,cm above the worktop at completion. Returning the cup to the table is not required for success. \\
\addlinespace[5pt]
FrameHang & \noncollabphase{\textbf{$p_1$ Grasp:}} Both robots hold the frame.
\collabphase{\textbf{$p_2$ Lift:}} They hold it together at least 10\,cm above its spawn height.
\collabphase{\textbf{$p_3$ Hang:}} The peg passes through the frame's loop while both robots hold the frame, which then remains threaded, stable, and released. \\
\addlinespace[5pt]
CoPouring & \collabphase{\textbf{$p_1$ Joint grasp and lift:}} Both robots hold the tub at least 3\,cm above its stack.
\collabphase{\textbf{$p_2$ Tip:}} They hold it together at a tilt of at least $25^\circ$.
\collabphase{\textbf{$p_3$ Pour:}} All apples are inside the receiving crate, with both robots still holding the tub. \\
\addlinespace[5pt]
TrashCollection & \noncollabphase{\textbf{$p_1$ Lift bin:}} A holds the bin at least 2\,cm off the floor.
\noncollabphase{\textbf{$p_2$ Deliver bin:}} A holds it aloft within 25\,cm of the bench-side carry point, with its rim no more than 1\,cm above the bench top.
\collabphase{\textbf{$p_3$ Sweep:}} The bottle is inside the bin, with A still holding the bin. \\
\addlinespace[5pt]
CartService & \noncollabphase{\textbf{$p_1$ Park:}} The trolley center reaches within 35\,cm of the parking mark.
\noncollabphase{\textbf{$p_2$ Lift bottle:}} B lifts the bottle from the trolley positioned by A to at least 10\,cm above its spawn height.
\noncollabphase{\textbf{$p_3$ Serve:}} B carries it within 15\,cm of the goal in the horizontal plane, then releases it in a stable, upright placement within $\pm6$\,cm of the goal on each axis. \\
\addlinespace[5pt]
CoCarry & \collabphase{\textbf{$p_1$ Joint grasp and lift:}} Both robots hold the basket with its base at least 3\,cm above the shelf.
\collabphase{\textbf{$p_2$ Carry:}} Both hold it within the goal's $\pm15$\,cm horizontal bounds, with at least 1\,cm base clearance above the shelf.
\collabphase{\textbf{$p_3$ Release:}} Stable placement within those bounds and $\pm6$\,cm of the target height, with all hands off. \\
\addlinespace[5pt]
TableAlign & \collabphase{\textbf{$p_1$ Joint grasp and lift:}} Both robots hold the table at least 3\,cm above its standing height.
\collabphase{\textbf{$p_2$ Carry:}} Both hold it inside the goal region with every foot bar at least 1\,cm off the floor.
\collabphase{\textbf{$p_3$ Align:}} Released, stable placement with the table center 33.5--46.5\,cm in front of the chair, within $\pm13$\,cm laterally and $\pm4$\,cm of standing height, and square to the chair within $4^\circ$. \\
\addlinespace[5pt]
MoveHouse & \textbf{$p_1$ Lift and open:} B and C jointly lift every table foot at least 3\,cm off the floor, and A opens the door to at least 1.45\,rad ($\approx 83^\circ$) for 0.5\,s. These two milestones may occur in either order.
\textbf{$p_2$ Pass doorway:} Table and both carriers pass the doorway clearance plane together, with B and C holding and the feet off the floor.
\textbf{$p_3$ Place:} B and C carry the table within 15\,cm of the goal while holding it aloft, then release it stably with each foot within its 10\,cm-square mark and height within $\pm4$\,cm of standing height. \\
\addlinespace[5pt]
BigTable & \textbf{$p_1$ Three-robot grasp and lift:} All three robots hold the table with the pedestal foot at least 3\,cm off the floor.
\textbf{$p_2$ Carry:} All three hold it with the entire pedestal foot over the goal disc and at least 1\,cm floor clearance.
\textbf{$p_3$ Release:} The pedestal foot remains within the disc, with stable, upright, released placement within $\pm5$\,cm of standing height. \\
\end{longtable}
\endgroup

\subsection{Language Instructions}
\label{app:instructions}

Each task provides a joint instruction and one role-specific instruction per robot: three strings for a two-robot task and four strings for a three-robot task.
Language-conditioned baselines generally use the corresponding role-specific instruction.
In the multi-agent design study, GR00T N1.7 instead uses the joint instruction in the \emph{Global Observation} + \emph{Shared Policy} configuration.
Table~\ref{tab:app-lang} lists all instruction strings without abbreviation.

\begingroup
\setlength{\LTcapwidth}{\linewidth}
\setlength{\tabcolsep}{5pt}
\renewcommand{\arraystretch}{1.08}
\newcommand{\instructiontask}[1]{%
  \multicolumn{2}{@{}l@{}}{%
    \begingroup
    \setlength{\fboxsep}{0pt}%
    \colorbox{catrow}{%
      \makebox[\linewidth][l]{%
        \rule[-0.6ex]{0pt}{2.5ex}\hspace{3pt}\textbf{#1}%
      }%
    }%
    \endgroup
  }\\*[3pt]
}
\begin{longtable}{@{}>{\raggedright\arraybackslash}p{0.11\linewidth}>{\raggedright\arraybackslash}p{\dimexpr0.89\linewidth-10pt\relax}@{}}
\caption{\textbf{Language instructions provided by the benchmark.} Joint denotes the shared instruction. Robot A, B, and C denote the corresponding robot's instruction.}\label{tab:app-lang}\\
\toprule
\textbf{Role} & \textbf{Instruction} \\
\midrule
\endfirsthead
\caption[]{(continued) \textbf{Language instructions provided by the benchmark.} Joint denotes the shared instruction. Robot A, B, and C denote the corresponding robot's instruction.}\\
\toprule
\textbf{Role} & \textbf{Instruction} \\
\midrule
\endhead
\endfoot
\bottomrule
\endlastfoot
\instructiontask{Handover}
\textbf{Joint} & Pass the bottle directly from Robot A's right hand to Robot B's right hand, then place it on the target. \\*[4pt]
\textbf{Robot A} & Pick up the bottle directly in front of you with your right hand and hand it to your partner's right hand. \\*[4pt]
\textbf{Robot B} & Receive the bottle with your right hand, place it on the target directly in front of you, and release it. \\
\addlinespace[6pt]
\instructiontask{Pouring}
\textbf{Joint} & Pour the marble from the small cup into the glass the other robot is holding in the air. \\*[4pt]
\textbf{Robot A} & Take the squat drinking cup in front of you round its body, hold it over the mouth of your partner's glass and tip it until the marble drops into the glass, then set the cup back on the table. \\*[4pt]
\textbf{Robot B} & Take the tall glass in front of you round its body, lift it off the table, and hold it out upright and still under your partner's cup until the marble is in it. \\
\addlinespace[6pt]
\instructiontask{FrameHang}
\textbf{Joint} & Lift the framed painting off the floor together, hang the loop on its top rail on the wall peg, and let go once it is hanging square. \\*[4pt]
\textbf{Robot A} & Take the left edge of the frame's moulding with one hand, lift level with your partner, guide the loop onto the peg, and release once it is seated. \\*[4pt]
\textbf{Robot B} & Take the right edge of the frame's moulding with one hand, lift level with your partner, guide the loop onto the peg, and release once it is seated. \\
\addlinespace[6pt]
\instructiontask{CoPouring}
\textbf{Joint} & Lift the tub of apples together and pour all of them into the empty crate beside you. \\*[4pt]
\textbf{Robot A} & Take the bar at your end of the tub with one hand, lift it level with your partner, and tip it together into the empty crate on your left until all the apples are in it. \\*[4pt]
\textbf{Robot B} & Take the bar at your end of the tub with one hand, lift it level with your partner, and tip it together into the empty crate on your right until all the apples are in it. \\
\addlinespace[6pt]
\instructiontask{TrashCollection}
\textbf{Joint} & Bring the bin to the bench and sweep the bottle off the bench into it. \\*[4pt]
\textbf{Robot A} & Bend and take the tall bin in both hands, carry it upright straight ahead to the end of your partner's bench, and hold it there with its rim just below the bench top. \\*[4pt]
\textbf{Robot B} & Wait at the bench. When the bin is at the end of it, sweep the plastic bottle lying between your hands along the surface and over that end into the bin; it rolls, so push straight and keep it off the near edge. \\
\addlinespace[6pt]
\instructiontask{CartService}
\textbf{Joint} & Push the loaded service trolley down the aisle to the green mark on the floor, then take the wine off it and stand the bottle on the green patch on the table. \\*[4pt]
\textbf{Robot A} & Take the push bar of the trolley in front of you with both hands and push it straight down the aisle until it stands on the green rectangle painted on the floor. \\*[4pt]
\textbf{Robot B} & Walk up the aisle to meet the trolley and stand square in front of its near edge; take the wine bottle around the glass just above the steel ring it stands in, lift it straight up out of the ring, then step round to your right to the near edge of the table and stand it on the green patch. \\
\addlinespace[6pt]
\instructiontask{CoCarry}
\textbf{Joint} & Carry the laundry basket together and set it down level on the green patch on the far shelf. \\*[4pt]
\textbf{Robot A} & Hold your side of the basket with both hands and side-step to your right, keeping it level, until it rests on the green patch. \\*[4pt]
\textbf{Robot B} & Hold your side of the basket with both hands and side-step to your left, keeping it level, until it rests on the green patch. \\
\addlinespace[6pt]
\instructiontask{TableAlign}
\textbf{Joint} & Carry the table to the chair together and set it down square to it. \\*[4pt]
\textbf{Robot A} & Take your end of the table with both hands, lift level with your partner, and side-step to your right down the room, turning the table a quarter circle together as you go, until it stands on the green patch with the chair tucked into its long edge. \\*[4pt]
\textbf{Robot B} & Take your end of the table with both hands, lift level with your partner, and side-step to your left down the room, turning the table a quarter circle together as you go, until it stands on the green patch with the chair tucked into its long edge. \\
\addlinespace[6pt]
\instructiontask{MoveHouse}
\textbf{Joint} & One robot opens the door and stands clear while the other two carry the table out of the bedroom, through the doorway, and into the living room. \\*[4pt]
\textbf{Robot A} & Walk to the door, take the lever, push the door wide open, and stand out of the lane so your partners and their table can pass. \\*[4pt]
\textbf{Robot B} & Take the two legs at your end of the table, one in each hand, lift level with your partner, and lead it backwards through the open doorway until its four legs stand on the green squares. \\*[4pt]
\textbf{Robot C} & Take the two legs at your end of the table, one in each hand, lift level with your partner, and follow them forwards through the open doorway until its four legs stand on the green squares. \\
\addlinespace[6pt]
\instructiontask{BigTable}
\textbf{Joint} & All three robots lift the round table together and carry it to the marked spot. \\*[4pt]
\textbf{Robot A} & Take the edge of the table in front of you with both hands, lift on the count, and walk it straight forward to the green disc, keeping it level. \\*[4pt]
\textbf{Robot B} & Take the edge of the table in front of you with both hands, lift on the count, and walk it backwards and to your right to the green disc, keeping it level. \\*[4pt]
\textbf{Robot C} & Take the edge of the table in front of you with both hands, lift on the count, and walk it backwards and to your left to the green disc, keeping it level. \\
\end{longtable}
\endgroup

\subsection{Demonstration Split and Usage}
\label{app:data-split}

Each task contains 60 demonstrations, of which the first 50 in collection order are used for training and the remaining 10 for validation. The same split is used across all baselines. Validation demonstrations are used only to monitor validation loss during training and are excluded from gradient updates.

\section{Experimental Setup Details}
\label{app:eval}

\subsection{Policy Inputs}
\label{app:baseline-inputs}

ACT and DP are trained separately for each task and robot.
The VLA and WAM policies instead share parameters across robot roles and tasks, with separate training on the two- and three-humanoid task suites.
ACT, DP, $\pi_{0.5}$, GR00T N1.7, and Fast-WAM condition on the controlled robot's egocentric RGB image and $43$ joint positions.
$\Psi_0$ instead uses its native $32$-D state, which consists of the $31$ arm, hand, and waist joint positions and the most recent base-height command, without the leg joints.
LatentToM jointly trains two policies, each using its own egocentric image and proprioception together with a shared third-person RGB view, while the policies execute independently.
GauDP jointly processes the agents' egocentric views and concatenated proprioception to predict their joint actions.
Its Gaussian encoder additionally uses recorded depth and camera geometry during training, but these signals are not required at policy inference.
Thus, LatentToM and GauDP have additional observation or training information that must be accounted for when comparing them with the egocentric-only baselines.

\subsection{Training Configurations}
\label{app:baseline-training}

Table~\ref{tab:app-hp} summarizes the training recipes. Batch sizes are global
batch sizes, including gradient accumulation where used. Epoch-based budgets
are reported as epochs because their optimizer-step counts depend on the task's
trajectory lengths. 
For evaluation, we use the last training checkpoint.

\begin{table}[!ht]
  \centering
  \caption{\textbf{Baseline training configurations.} Budgets refer to policy
  training; GauDP's Gaussian encoder is trained separately as described below.
  Peak LR denotes the maximum learning rate. Band shading follows
  Table~\ref{tab:main-compact}: blue for the families trained separately per
  task, green for the VLA and WAM families trained jointly across tasks.}
  \label{tab:app-hp}
  \small
  \setlength{\tabcolsep}{4pt}
  \renewcommand{\arraystretch}{1.15}
  \begin{tabularx}{\linewidth}{@{}l>{\raggedright\arraybackslash}Xccc@{}}
    \toprule
    \textbf{Policy} & \textbf{Network initialization and training}
      & \textbf{Batch} & \textbf{Budget} & \textbf{Peak LR} \\
    \midrule
    \famrow{5}{Standard IL policies}
    ACT & Pretrained vision encoder; encoder finetuning and transformer training
      & $256$ & $600$ epochs & $10^{-4}$ \\
    DP & From scratch; full policy training
      & $256$ & $600$ epochs & $10^{-4}$ \\
    \addlinespace[3pt]
    \famrow{5}{Multi-agent IL policies}
    LatentToM & From scratch; joint training of two policies
      & $32$ & $1{,}000$ epochs & $10^{-4}$ \\
    GauDP & Pretrained Gaussian encoder; encoder finetuning, then diffusion policy training
      & $128$ & $150$ epochs & $10^{-4}$ \\
    \addlinespace[3pt]
    \famrowmt{5}{Vision-Language-Action (VLA) models}
    $\pi_{0.5}$ & Pretrained VLA; LoRA finetuning
      & $32$ & $40{,}000$ steps & $2.5\!\times\!10^{-5}$ \\
    GR00T N1.7 & Pretrained VLA; LoRA finetuning
      & $32$ & $40{,}000$ steps & $2.5\!\times\!10^{-5}$ \\
    $\Psi_0$ & Pretrained VLA; LoRA finetuning
      & $32$ & $40{,}000$ steps & $2.5\!\times\!10^{-5}$ \\
    \addlinespace[3pt]
    \famrowmt{5}{World Action Models (WAMs)}
    Fast-WAM & Pretrained video backbone; LoRA finetuning
      & $32$ & $40{,}000$ steps & $10^{-4}$ \\
    \bottomrule
  \end{tabularx}
\end{table}

\noindent\textbf{Standard IL policies.}
ACT uses an ImageNet-pretrained ResNet-18~\citep{he2016deep} as its vision encoder and predicts actions in chunks of $50$ steps.
Training uses a constant learning rate, and temporal aggregation is disabled at inference.
DP uses a ResNet-18 vision encoder trained from scratch and a DDPM~\citep{ho2020denoising} with $100$ diffusion steps during both training and inference.
It predicts $40$ actions and executes the first $20$ before replanning.
Images are resized to $320\times240$, with $288\times216$ random crops during training and center crops at inference.
Its learning rate follows a cosine schedule after $500$ warmup steps.

\noindent\textbf{Multi-agent IL policies.}
LatentToM uses a sheaf-consistency loss with weight $0.5$ to couple its two policies during training.
Its diffusion scheduler is DDIM~\citep{song2021denoising}, with $100$ training and inference steps.
GauDP first finetunes a shared NoPoSplat~\citep{ye2025no} pretrained encoder on the two- and three-humanoid task suites for up to $10$ and $8$ epochs, respectively, using batch size $8$.
The encoder is then frozen, Gaussian features are cached, and a separate DDPM policy is trained for each task.
Both methods use EMA and a single observation step, predicting 40 actions and executing 20 at each inference step.
During training, both add Gaussian proprioceptive noise with standard deviation $0.02$ times half of each dimension's demonstration range.

\noindent\textbf{Vision-Language-Action (VLA) models.}
All three VLAs are fine-tuned with the same recipe.
We apply LoRA to the LLM backbone (rank $16$, $\alpha{=}16$) and to the action head (rank $32$, $\alpha{=}32$) of all three models, while the vision encoder is fully fine-tuned and the remaining backbone weights are frozen.
We use the AdamW optimizer~\citep{loshchilov2019decoupled} with momentum coefficients $(\beta_1,\beta_2)=(0.9,0.95)$ and weight decay $10^{-10}$.
Gradients are clipped to a maximum norm of $1.0$.
After $1{,}000$ warmup steps, the learning rate follows a cosine decay from $2.5\times10^{-5}$ to $2.5\times10^{-6}$.
Each policy conditions on a single-frame observation and predicts an action chunk that is executed in full before replanning: $50$ steps for $\pi_{0.5}$, $40$ for GR00T N1.7, and $30$ for $\Psi_0$.

\noindent\textbf{World Action Model (WAM).}
Fast-WAM initializes its video expert from Wan2.2~\citep{wan2025} and fine-tunes it using LoRA (rank 128, $\alpha=64$), while fully training the action expert.
It uses a $32$-action prediction horizon and a cosine learning-rate schedule.

\subsection{Rollout Evaluation}
\label{app:baseline-evaluation}

The benchmark evaluation uses $100$ rollouts per policy and task, with matched environment seeds $1$--$100$.
Initial object and target poses are sampled according to the ranges specified in Appendix~\ref{app:task-scenes}.
At $50$\,Hz, the episode limits are $1{,}000$ control steps ($20$\,s) for most tasks, $1{,}100$ ($22$\,s) for TableAlign, $1{,}500$ ($30$\,s) for MoveHouse, and $2{,}000$ ($40$\,s) for CartService.
Each rollout records overall success and the cumulative phase outcomes defined in Appendix~\ref{app:task-phases}.

\subsection{Agentic Robot Policy Evaluation}
\label{app:astra-evaluation}

\noindent\textbf{Policy and observations.}
We evaluate GPT-6 Astra~\citep{openai2026astra} zero-shot, without training on \bench data or replaying demonstrations. 
A fresh Codex session is created for each episode,
with high reasoning effort and a task-specific instruction describing the
robots' roles and the control interface. 

One centralized policy controls both humanoids, receiving their egocentric RGB images and proprioception, including $43$ joint positions per robot and measured wrist poses.
Depth is accessed only through pixel-wise queries at selected image locations.
Wrist poses and camera calibration are expressed in each robot's own pelvis frame.
No simulator-provided object poses are exposed.

\noindent\textbf{Control and feedback.}
Within each episode, Astra repeatedly inspects both robots' egocentric images and state, outputs commands for the two robots, and examines the resulting observations and execution feedback before deciding its next action. 

It interacts with the simulator by calling Python helper functions that provide robots' egocentric observations, accept commands for robots, and return execution feedback through local JSON files. A single Codex session maintains the conversation history across these interactions, allowing Astra to revise its plan based on earlier actions and feedback.

Astra directly outputs seven finger joint-angle targets per hand, body-frame planar base velocities, yaw rate, and a pelvis-height target.
For wrist position and orientation, Astra outputs wrist pose targets rather than joint-angle targets.
Pink~\citep{caron2026pink} solves the coordinated arm and waist motion from the target pose.
A Python helper is provided to aid conversion from target pose into joints, using Pink.

Wrist targets are transformed from the observed pelvis frame and held fixed during execution,
so walking does not automatically carry the hands along; Astra needs to move the wrist targets accordingly.
Measured wrist errors and joint-limit violations are given as feedback to support replanning and within-episode recovery.

\noindent\textbf{Evaluation protocol.}
Actions execute in chunks of at most $25$ control steps at $50$\,Hz, excluding policy reasoning time. 
Each episode allows $6{,}000$ steps ($120$\,s)
including a $300$-step initialization hold, and a $30$-minute policy wall-clock
budget. Environment steps are not computed during policy reasoning.
Each task is evaluated over $10$ episodes with environment seeds $0$--$9$.

\begin{table}[!ht]
  \caption{\textbf{Astra phase-wise success rates (\%) on tasks with two humanoids and low base movement.}
  Phases are defined in Appendix~\ref{app:tasks}; $p_3$ is full-task success.}
  \label{tab:app-astra}
  \centering
  \footnotesize
  \setlength{\tabcolsep}{2.6pt}
  \renewcommand{\arraystretch}{1.05}
  \begin{tabularx}{\linewidth}{@{}l*{12}{C}@{}}
    \toprule
    \textbf{Method} & \multicolumn{3}{c}{\textbf{Handover}} & \multicolumn{3}{c}{\textbf{Pouring}} & \multicolumn{3}{c}{\textbf{FrameHang}} & \multicolumn{3}{c}{\textbf{CoPouring}} \\
    \cmidrule(lr){2-4}\cmidrule(lr){5-7}\cmidrule(lr){8-10}\cmidrule(lr){11-13}
      & \noncollabphase{$p_1$} & \collabphase{$p_2$} & \noncollabphase{$p_3$} & \noncollabphase{$p_1$} & \noncollabphase{$p_2$} & \collabphase{$p_3$} & \noncollabphase{$p_1$} & \collabphase{$p_2$} & \collabphase{$p_3$} & \collabphase{$p_1$} & \collabphase{$p_2$} & \collabphase{$p_3$} \\
    \midrule
    GPT-6 Astra & 10 & 0 & 0 & 90 & 80 & 10 & 100 & 60 & 0 & 100 & 70 & 70 \\
    \bottomrule
  \end{tabularx}
\end{table}

\begin{table}[!ht]
  \caption{\textbf{Astra phase-wise success rates (\%) on tasks with two humanoids and high base movement.}
  Phases are defined in Appendix~\ref{app:tasks}; $p_3$ is full-task success.}
  \label{tab:app-astra-highloco}
  \centering
  \footnotesize
  \setlength{\tabcolsep}{2.6pt}
  \renewcommand{\arraystretch}{1.05}
  \begin{tabularx}{\linewidth}{@{}l*{12}{C}@{}}
    \toprule
    \textbf{Method} & \multicolumn{3}{c}{\textbf{TrashCollection}} & \multicolumn{3}{c}{\textbf{CartService}} & \multicolumn{3}{c}{\textbf{CoCarry}} & \multicolumn{3}{c}{\textbf{TableAlign}} \\
    \cmidrule(lr){2-4}\cmidrule(lr){5-7}\cmidrule(lr){8-10}\cmidrule(lr){11-13}
      & \noncollabphase{$p_1$} & \noncollabphase{$p_2$} & \collabphase{$p_3$} & \noncollabphase{$p_1$} & 
      \noncollabphase{$p_2$} & \noncollabphase{$p_3$} & \collabphase{$p_1$} & \collabphase{$p_2$} & \collabphase{$p_3$} & \collabphase{$p_1$} & \collabphase{$p_2$} & \collabphase{$p_3$} \\
    \midrule
    GPT-6 Astra & 90 & 0 & 0 & 70 & 0 & 0 & 40 & 0 & 0 & 80 & 10 & 0 \\
    \bottomrule
  \end{tabularx}
\end{table}

\noindent\textbf{Results.}
Tables~\ref{tab:app-astra} and~\ref{tab:app-astra-highloco} report phase-wise success rates for the 8 two-humanoid tasks.
Astra succeeds in 70\% of CoPouring and 10\% of Pouring, but records no successful episodes on the remaining tasks.

\input{tables/phasewise-tables}

\clearpage
\section{Collaboration Failure Cases}
\label{app:failures}

Figure~\ref{fig:task-failures} presents representative failure cases across all ten tasks, providing a qualitative view of how multi-humanoid collaboration can break down in \bench.

\begingroup
\setlength{\intextsep}{6pt}
\begin{figure}[!ht]
\centering
\newcommand{\failfig}[1]{figures/task_failure_figures/#1/#1_fail}
\failsetlengths
\setlength{\failstripw}{0.46\linewidth}
\setlength{\failimgw}{\dimexpr(\failstripw-\failimgsep)/2\relax}
\failcell{(a) Handover}{\failfig{handover}_1.jpg}{\failfig{handover}_2.jpg}\hfill
\failcell{(b) Pouring}{\failfig{pouring}_1.jpg}{\failfig{pouring}_2.jpg}\\[5pt]
\failcell{(c) FrameHang}{\failfig{framehang}_1.jpg}{\failfig{framehang}_2.jpg}\hfill
\failcell{(d) CoPouring}{\failfig{copouring}_1.jpg}{\failfig{copouring}_2.jpg}\\[5pt]
\failcell{(e) TrashCollection}{\failfig{trashcollection}_1.jpg}{\failfig{trashcollection}_2.jpg}\hfill
\failcell{(f) CartService}{\failfig{cartservice}_1.jpg}{\failfig{cartservice}_2.jpg}\\[5pt]
\failcell{(g) CoCarry}{\failfig{cocarry}_1.jpg}{\failfig{cocarry}_2.jpg}\hfill
\failcell{(h) TableAlign}{\failfig{tablealign}_1.jpg}{\failfig{tablealign}_2.jpg}\\[5pt]
\failcell{(i) MoveHouse}{\failfig{movehouse}_1.jpg}{\failfig{movehouse}_2.jpg}\hfill
\failcell{(j) BigTable}{\failfig{bigtable}_1.jpg}{\failfig{bigtable}_2.jpg}
\caption{\textbf{Representative collaboration failures across the \bench task suite.}
Each pair shows two moments of one GR00T N1.7 rollout, earlier on the left.
\textbf{(a) Handover:} the giver lets go before the receiver's hand closes on the bottle, which drops between the two hands.
\textbf{(b) Pouring:} the cup is tipped while the partner's glass is not beneath it, so the marble misses the glass.
\textbf{(c) FrameHang:} the robots lift the frame at mismatched heights and angles, so its loop misses the peg and the frame slips.
\textbf{(d) CoPouring:} the robots fail to synchronize their grasps, leaving one handle ungrasped and causing the apples to spill outside the receiving crate.
\textbf{(e) TrashCollection:} the bottle is swept off the laboratory table but misses the bin and falls to the floor.
\textbf{(f) CartService:} the receiving robot carries the bottle to the goal table but walks past it instead of placing it.
\textbf{(g) CoCarry:} one robot loses its grip on the basket, unbalancing the load and causing it to fall.
\textbf{(h) TableAlign:} one robot loses its grip during transport, and the table falls outside the target region.
\textbf{(i) MoveHouse:} the two robot carriers fail to coordinate their motion, causing the table to deviate from the doorway path and get stuck against the door frame.
\textbf{(j) BigTable:} during transport, one robot loses its grip on the table, which tilts and never reaches the goal.}
\label{fig:task-failures}
\end{figure}
\endgroup

\end{document}

%% file: tables/main-results-table.tex
\begin{table}[t]
    \caption{\textbf{Main benchmark results.}
    Final task success rate (\%) on the \bench task suite, evaluated from the beginning of each task. Blue and green rows denote per-task and multi-task training, respectively.
    Best result per task and best average are in \textbf{bold}. The shaded \textbf{Avg.} columns average across tasks and \emph{Task avg.} across policies.
    Phase-wise results are provided in Appendix~\ref{app:phasewise}.}
  \label{tab:main-compact}
  \definecolor{avgcellbg}{gray}{0.93}%
  \centering
  \footnotesize
  \setlength{\tabcolsep}{3.4pt}
  \renewcommand{\arraystretch}{1.05}
  \begin{tabular*}{\linewidth}{@{\extracolsep{\fill}}l*{8}{w{c}{1.8em}} >{\columncolor{avgcellbg}[4pt][4pt]}w{c}{1.8em}@{\hspace{5pt}}!{\vrule width 0.4pt}@{\hspace{5pt}}w{c}{1.8em}w{c}{1.8em} >{\columncolor{avgcellbg}[4pt][4pt]}w{c}{1.8em}@{\hspace{4pt}}}
    \toprule
      & \multicolumn{9}{c}{\textbf{2 Humanoids}}
      & \multicolumn{3}{c}{\hspace*{-6.2pt}\textbf{3 Humanoids}} \\
    \cmidrule(lr){2-10}\cmidrule(l{-6.5pt}){11-13}
    \textbf{Method}
      & \rot{Handover} & \rot{Pouring} & \rot{FrameHang} & \rot{CoPouring} & \rot{TrashCollection} & \rot{CartService} & \rot{CoCarry} & \rot{TableAlign}
      & \rot{\textbf{Avg.}}
      & \rot{MoveHouse} & \rot{BigTable}
      & \rot{\textbf{Avg.}} \\
    \midrule
    \famrow{13}{Standard IL policies}
    ACT              & \best{65} & 1 & \best{44} & 9 & 1 & 3 & \best{61} & \best{39} & \best{27.9} & 9 & \best{80} & \best{44.5} \\
    DP               & 30 & 0 & 26 & 7 & 1 & 3 & 12 & 4 & 10.4 & 4 & 14 & 9.0 \\
    \addlinespace[2pt]
    \famrow{13}{Multi-agent IL policies}
    LatentToM$^{\dagger}$ & 5 & 0 & 2 & 1 & 0 & 0 & 31 & 3 & 5.3 & \na & \na & \na \\
    GauDP            & 52 & 0 & 29 & 0 & 1 & 0 & 9 & 16 & 13.4 & 2 & 54 & 28.0 \\
    \addlinespace[2pt]
    \famrowmt{13}{Vision-Language-Action (VLA) models}
    $\pi_{0.5}$      & 2 & 0 & 27 & 0 & 0 & \best{4} & 1 & 2 & 4.5 & 3 & 37 & 20.0 \\
    GR00T N1.7       & 27 & 0 & \best{44} & 0 & 1 & 0 & 2 & 1 & 9.4 & 0 & 5 & 2.5 \\
    $\Psi_0$         & 0 & 0 & 5 & 0 & \best{3} & 0 & 0 & 0 & 1.0 & 0 & 19 & 9.5 \\
    \addlinespace[2pt]
    \famrowmt{13}{World Action Models (WAMs)}
    Fast-WAM         & 2 & \best{7} & 28 & \best{11} & 1 & 0 & 31 & 17 & 12.1 & \best{20} & 69 & \best{44.5} \\
    \midrule
    \emph{Task avg.}  & 22.9 & 1.0 & 25.6 & 3.5 & 1.0 & 1.3 & 18.4 & 10.3 & \multicolumn{1}{c@{\hspace{5pt}}!{\vrule width 0.4pt}@{\hspace{5pt}}}{}
      & 5.4 & 39.7 & \multicolumn{1}{c@{\hspace{4pt}}}{} \\
    \bottomrule
  \end{tabular*}
  \par\vspace{3pt}
  \begin{minipage}{\linewidth}
    \footnotesize\raggedright
    $^{\dagger}$ We evaluate LatentToM only on tasks with two robots, following
    the original paper's experimental setup.
  \end{minipage}
\end{table}

%% file: figures/fig-collab-cond-gap.tex
\DeclareFontShape{OT1}{cmr}{m}{n}{%
  <-6>cmr5<6-7>cmr6<7-8>cmr7<8-9>cmr8<9-10>cmr9<10-12>cmr10<12-17>cmr12<17->cmr17}{}
\DeclareFontShape{OML}{cmm}{m}{it}{%
  <-6>cmmi5<6-7>cmmi6<7-8>cmmi7<8-9>cmmi8<9-10>cmmi9<10-12>cmmi10<12->cmmi12}{}
\DeclareFontShape{OMS}{cmsy}{m}{n}{%
  <-6>cmsy5<6-7>cmsy6<7-8>cmsy7<8-9>cmsy8<9->cmsy10}{}

\makeatletter
\newif\ifccndpgfplots
\@ifpackageloaded{pgfplots}{\ccndpgfplotstrue}{\ccndpgfplotsfalse}
\makeatother
\ifccndpgfplots

\definecolor{ccndNon}{HTML}{79B1CD}
\definecolor{ccndCol}{HTML}{B2D4A6}
\definecolor{ccndDrop}{HTML}{DC0000}
\definecolor{ccndGrid}{HTML}{D6D6D3}
\definecolor{ccndSub}{HTML}{55554F}
\definecolor{ccndFam}{HTML}{6F6F6A}

\pgfplotsset{
  ccnd panel/.style={
    width=12.8cm, height=3.0cm, scale only axis,
    ybar, bar width=11pt,
    ymin=0, ymax=118, ytick={0,20,40,60,80,100},
    yticklabels={0,20,40,60,80,100},
    xtick=data, symbolic x coords={act,dp,ltom,gaudp,pi,gr,psi,wam},
    xtick style={draw=none},
    enlarge x limits=0.08,
    axis x line*=bottom, axis y line*=left,
    x axis line style={draw=black, line width=0.6pt},
    y axis line style={draw=black, line width=0.6pt},
    ymajorgrids, grid style={draw=ccndGrid, line width=0.5pt},
    tick style={draw=black, line width=0.5pt}, tick align=outside,
    clip=false,
    xticklabel style={font=\ttfamily\fontsize{7.0}{8.0}\selectfont,
                      align=center, inner sep=2.6pt},
    yticklabel style={font=\ttfamily\fontsize{7.2}{8.2}\selectfont, xshift=1pt},
    ylabel style={font=\ttfamily\fontsize{8.0}{9.0}\selectfont, yshift=-4pt},
  },
  ccnd shadow/.style={fill=black, draw=black, line width=0.4pt,
                      xshift=1.2pt, yshift=-1.2pt, forget plot},
  ccnd bar/.style={draw=black, line width=0.5pt, forget plot},
}

\newcommand{\ccndval}[3]{%
  \node[anchor=south, inner sep=0pt, xshift={#2+0.6pt}, yshift=2.0pt,
        font=\ttfamily\fontsize{6.8}{7.6}\selectfont, color=black,
        text depth=0pt]
    at (axis cs:#1,#3) {#3};}

\newcommand{\ccnddropsize}{9.0}
\newcommand{\ccnddrop}[2]{%
  \node[anchor=south, inner sep=0pt, xshift=0.6pt,
        font=\ttfamily\fontsize{\ccnddropsize}{\ccnddropsize}\selectfont, color=ccndDrop,
        text depth=0pt]
    at (axis cs:#1,111) {#2\%};}
\newcommand{\ccnddropmax}[2]{%
  \node[anchor=south, inner sep=0pt, xshift=0.6pt,
        font=\ttfamily\bfseries\fontsize{\ccnddropsize}{\ccnddropsize}\selectfont, color=ccndDrop,
        text depth=0pt]
    at (axis cs:#1,111) {#2\%};}

\newcommand{\ccndfam}[3]{%
  \draw[draw=ccndFam, line width=0.5pt]
    ([xshift=-18pt,yshift=-13.0pt]axis cs:#1,0) --
    ([xshift=-18pt,yshift=-15.6pt]axis cs:#1,0);
  \draw[draw=ccndFam, line width=0.5pt]
    ([xshift=-18pt,yshift=-15.6pt]axis cs:#1,0) --
    ([xshift=18pt,yshift=-15.6pt]axis cs:#2,0)
    node[midway, anchor=north, yshift=-1.0pt, color=ccndFam,
         font=\ttfamily\fontsize{6.6}{7.4}\selectfont] {#3};
  \draw[draw=ccndFam, line width=0.5pt]
    ([xshift=18pt,yshift=-15.6pt]axis cs:#2,0) --
    ([xshift=18pt,yshift=-13.0pt]axis cs:#2,0);}

\newcommand{\ccndswatch}[1]{%
  \raisebox{0.5\fontcharht\font`N}{\tikz[baseline={(0,3.1pt)}]{\fill[#1] (0,0) rectangle (8.4pt,6.2pt);
    \draw[black, line width=0.5pt] (0,0) rectangle (8.4pt,6.2pt);}}}

\begin{figure}[t]
  \centering
  \begin{tikzpicture}
    \begin{axis}[
      ccnd panel,
      xticklabels={ACT, DP, LatentToM, GauDP, $\pi_{0.5}$, GR00T N1.7,
                   $\Psi_0$, Fast-WAM},
      ylabel={Phase Success Rate (\%)},
    ]
      \addplot[ccnd shadow, bar shift=-7pt]
        coordinates {(act,86.2) (dp,72.0) (ltom,63.4) (gaudp,69.8)
                     (pi,76.6) (gr,73.4) (psi,54.8) (wam,76.8)};
      \addplot[ccnd shadow, bar shift=7pt]
        coordinates {(act,72.1) (dp,65.0) (ltom,56.1) (gaudp,60.6)
                     (pi,52.4) (gr,53.0) (psi,46.4) (wam,62.7)};
      \addplot[ccnd bar, bar shift=-7pt, fill=ccndNon]
        coordinates {(act,86.2) (dp,72.0) (ltom,63.4) (gaudp,69.8)
                     (pi,76.6) (gr,73.4) (psi,54.8) (wam,76.8)};
      \addplot[ccnd bar, bar shift=7pt, fill=ccndCol]
        coordinates {(act,72.1) (dp,65.0) (ltom,56.1) (gaudp,60.6)
                     (pi,52.4) (gr,53.0) (psi,46.4) (wam,62.7)};

      \ccndval{act}{-7pt}{86.2}   \ccndval{act}{7pt}{72.1}
      \ccndval{dp}{-7pt}{72.0}    \ccndval{dp}{7pt}{65.0}
      \ccndval{ltom}{-7pt}{63.4}  \ccndval{ltom}{7pt}{56.1}
      \ccndval{gaudp}{-7pt}{69.8} \ccndval{gaudp}{7pt}{60.6}
      \ccndval{pi}{-7pt}{76.6}    \ccndval{pi}{7pt}{52.4}
      \ccndval{gr}{-7pt}{73.4}    \ccndval{gr}{7pt}{53.0}
      \ccndval{psi}{-7pt}{54.8}   \ccndval{psi}{7pt}{46.4}
      \ccndval{wam}{-7pt}{76.8}   \ccndval{wam}{7pt}{62.7}

      \ccnddrop{act}{-16.3}    \ccnddrop{dp}{-9.7}
      \ccnddrop{ltom}{-11.4}   \ccnddrop{gaudp}{-13.2}
      \ccnddropmax{pi}{-31.6}  \ccnddrop{gr}{-27.8}
      \ccnddrop{psi}{-15.3}    \ccnddrop{wam}{-18.3}

      \draw[draw=ccndFam, line width=0.6pt, dash pattern=on 2.5pt off 2pt]
        ([xshift=22.45pt]axis cs:gaudp,0) -- ([xshift=22.45pt]axis cs:gaudp,118);

      \ccndfam{act}{dp}{Standard IL}
      \ccndfam{ltom}{gaudp}{Multi-agent IL}
      \ccndfam{pi}{psi}{VLA}
      \ccndfam{wam}{wam}{WAM}
    \end{axis}

    \node[anchor=south west, draw=black, line width=0.5pt, fill=white,
          inner xsep=4pt, inner ysep=3pt,
          font=\ttfamily\fontsize{7.8}{8.8}\selectfont]
      at ([yshift=4pt]current bounding box.north west)
      {\ccndswatch{ccndNon}~Non-collaborative (NC)\quad
       \ccndswatch{ccndCol}~Collaborative (C)\quad
       {\color{ccndDrop}\fontsize{\ccnddropsize}{8.8}\selectfont -x\%}~Relative drop (C-NC)/NC};
  \end{tikzpicture}
  \par\vspace{-8pt}
    \caption{\textbf{Phase-level evaluation results.}
    Average success rate on \textcolor{noncollabphase}{\textbf{non-collaborative}} and
    \textcolor{collabphase}{\textbf{collaborative}} phases over the 8 tasks with two humanoids.
    Each phase is evaluated from its beginning state in each of the same 50 demonstrations. The red numbers indicate the relative drops.}
  \label{fig:collab-cond-gap}
\end{figure}

\else
\PackageWarningNoLine{cohub}{figures/fig-collab-cond-gap.tex needs pgfplots --
  add \string\usepackage{pgfplots} and \string\pgfplotsset{compat=1.18} to the
  preamble}%
\begin{figure}[t]
  \centering
  \fbox{\parbox[c][2.6cm][c]{0.90\linewidth}{\centering\small
    This figure needs \texttt{pgfplots}. Add
    \texttt{\string\usepackage\{pgfplots\}} and
    \texttt{\string\pgfplotsset\{compat=1.18\}} to the preamble.}}
  \caption{Per-phase success rate on collaborative versus non-collaborative
  phases (placeholder).}
  \label{fig:collab-cond-gap}
\end{figure}
\fi
\makeatother

%% file: tables/obs-act-table.tex
\begin{wraptable}{r}{0.5\textwidth}
  \vspace{-20pt}
  \caption{\textbf{Effect of observation scope and policy sharing.}
  Average task success rate (\%) over the 8 tasks with two humanoids across different observation scopes and policy-sharing strategies.}
  \label{tab:view-policy}
  \centering
  \footnotesize
  \setlength{\tabcolsep}{3pt}
  \renewcommand{\arraystretch}{1.05}
  \begin{tabular}{clccc}
    \toprule
      & & \multicolumn{2}{c}{\textbf{Standard IL}} & \multicolumn{1}{c}{\textbf{VLA}} \\
    \cmidrule(lr){3-4}\cmidrule(l){5-5}
    \textbf{Obs. Scope} & \textbf{Policy} & ACT & DP & GR00T N1.7 \\
    \midrule
    Local  & Separate & \textbf{27.9} & 10.4 & \textbf{14.5} \\
    Local  & Shared   & 23.9 & 9.5 & 9.4 \\
    Global & Separate & 25.6 & \textbf{19.1} & 12.3 \\
    Global & Shared   & 26.4 & 16.0 & 9.0 \\
    \bottomrule
  \end{tabular}
  \vspace{-8pt}
\end{wraptable}

%% file: figures/task-gallery.tex
\providecommand{\galleryimgtrim}[5]{%
  \adjustbox{trim={#2\width} {#3\height} {#4\width} {#5\height},clip,%
             width=\taskframew}{\includegraphics{#1}}}

\begin{figure}[p]
  \centering
  \setlength{\taskstripw}{0.84\linewidth}%
  \setlength{\taskimgsep}{0.015\linewidth}%
  \setlength{\taskframew}{\dimexpr(\taskstripw-2\taskimgsep)/3\relax}%
  \taskcell{(a) Handover}{%
    \taskimg{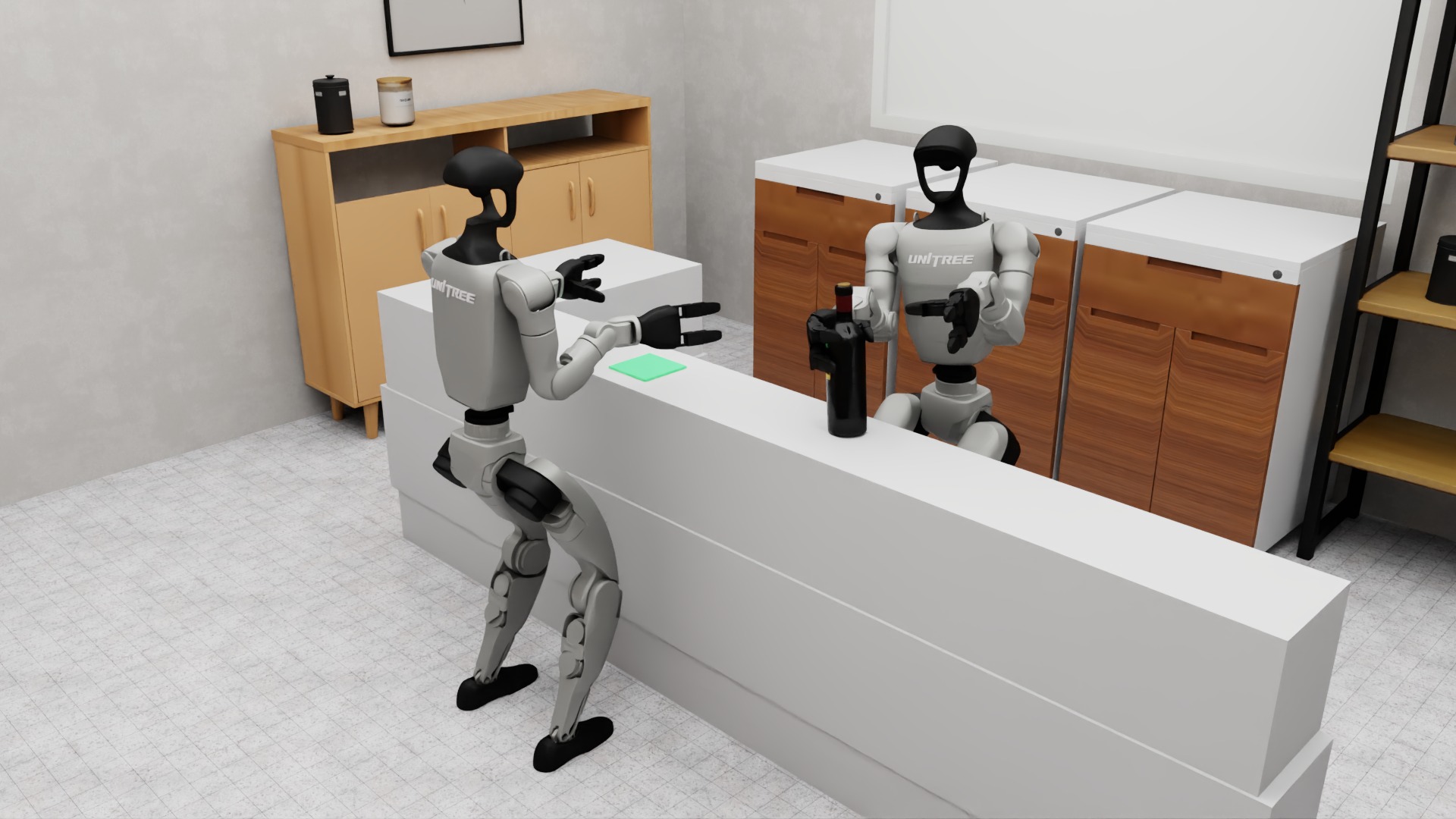}\hspace{\taskimgsep}%
    \taskimg{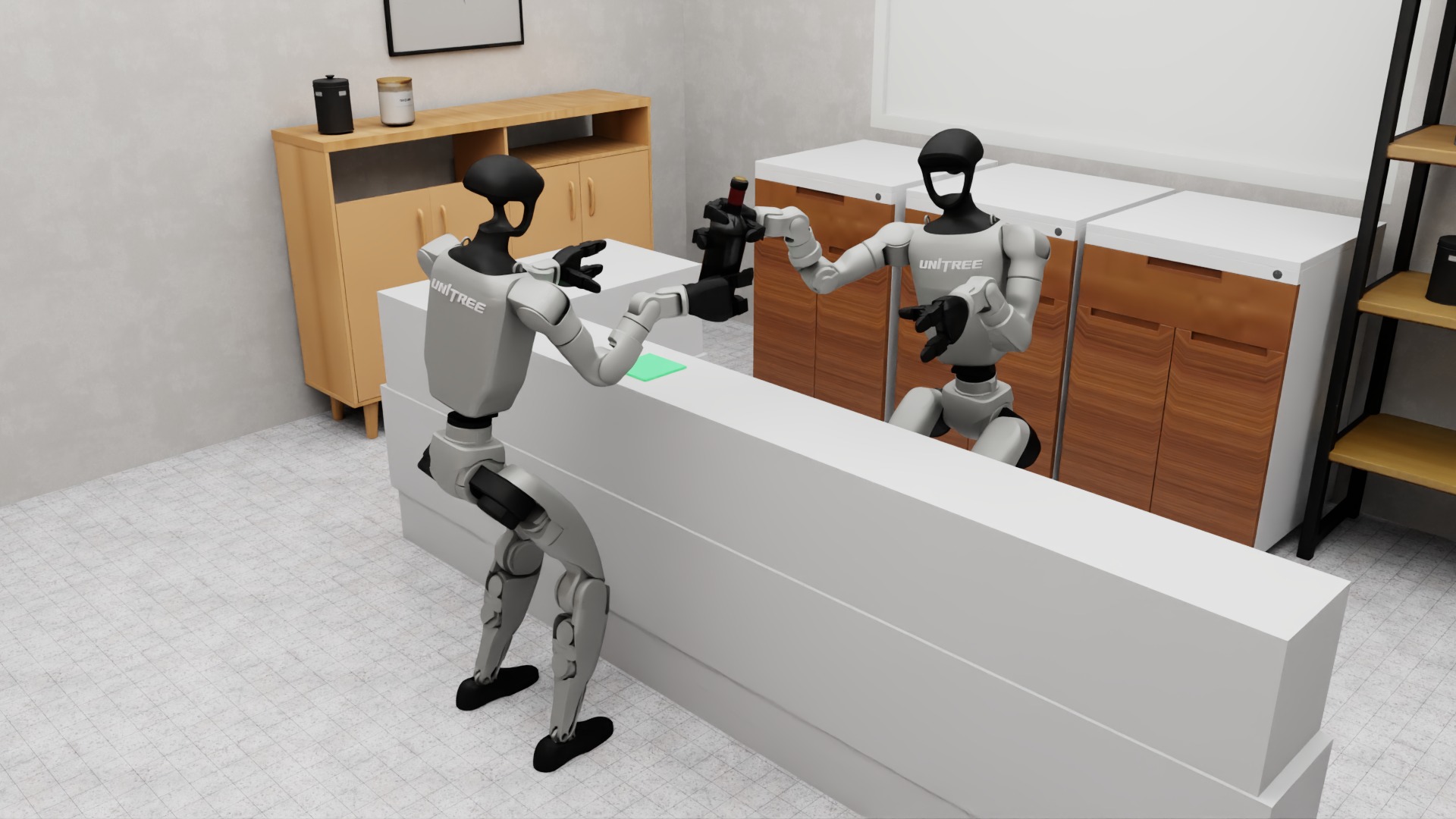}\hspace{\taskimgsep}%
    \taskimg{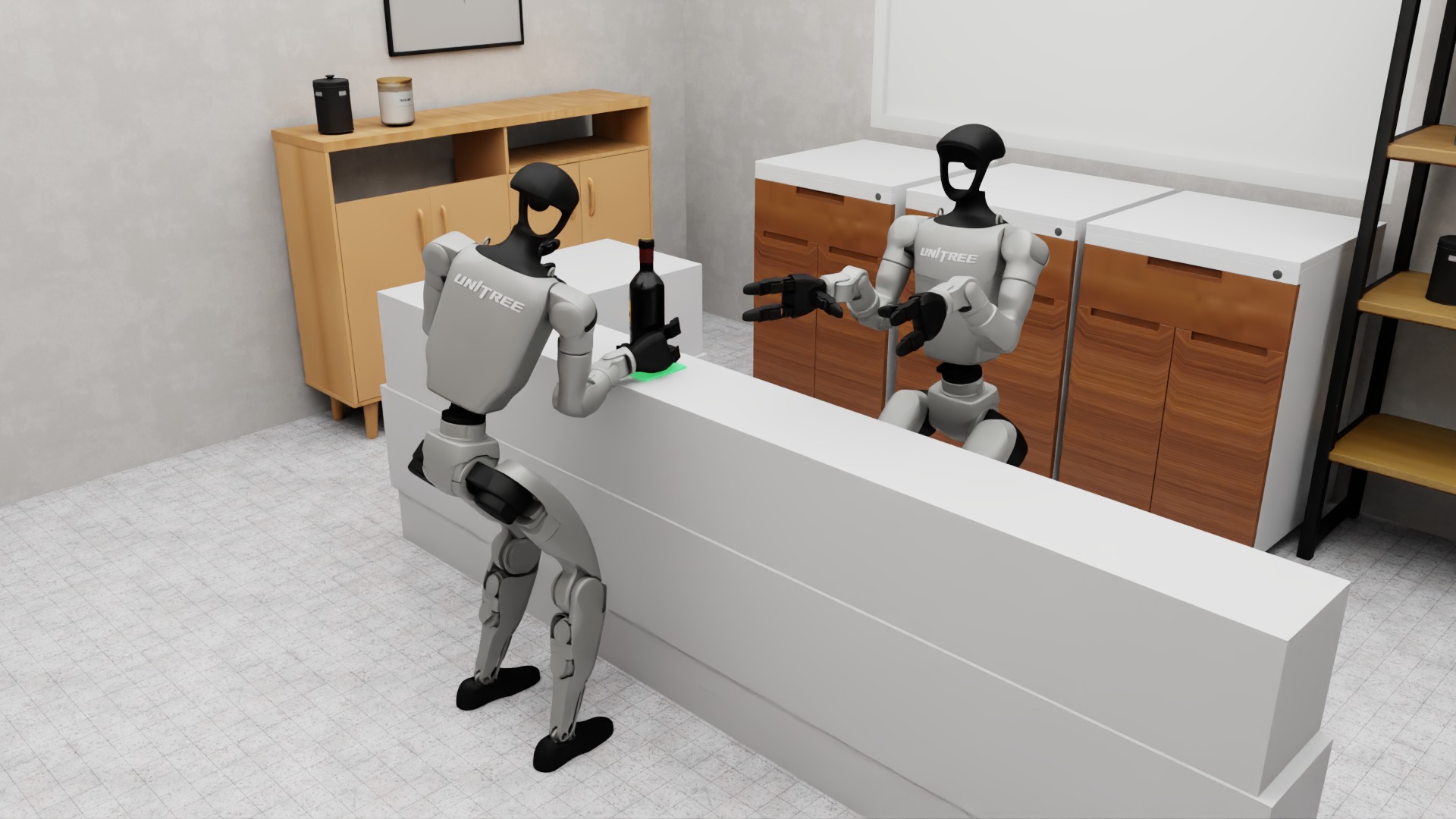}}\\[6pt]
  \taskcell{(b) Pouring}{%
    \taskimg{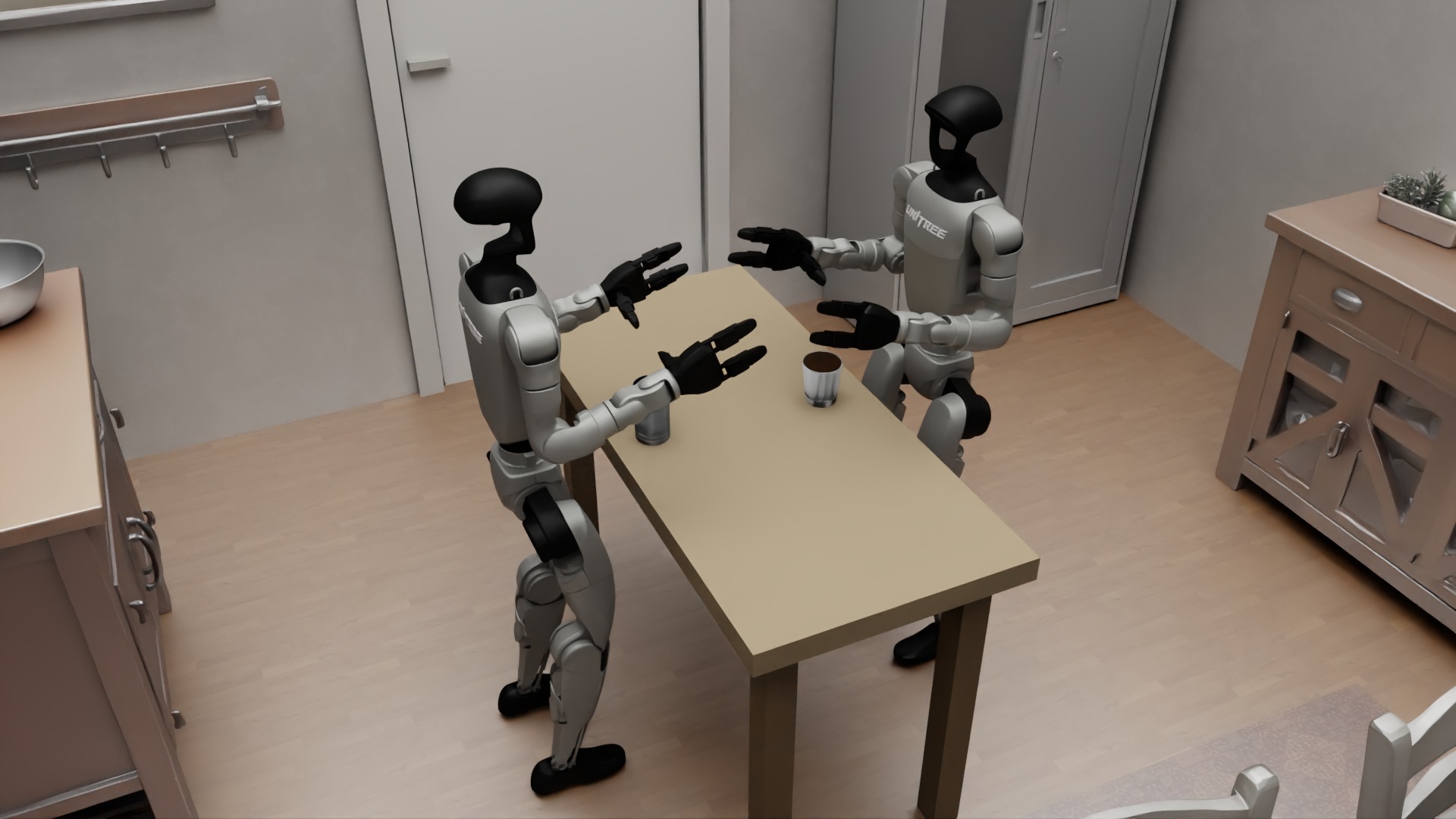}\hspace{\taskimgsep}%
    \taskimg{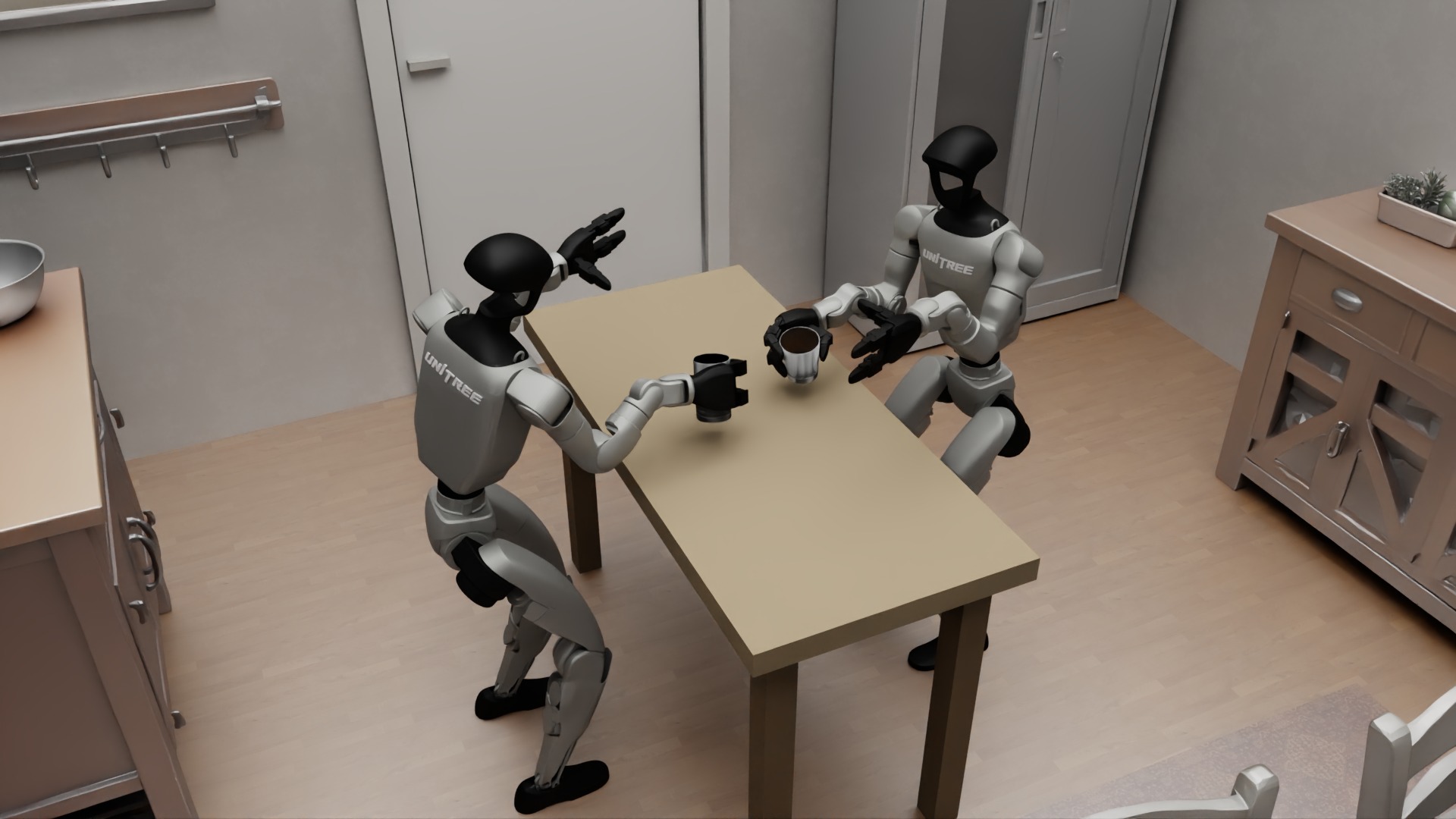}\hspace{\taskimgsep}%
    \taskimg{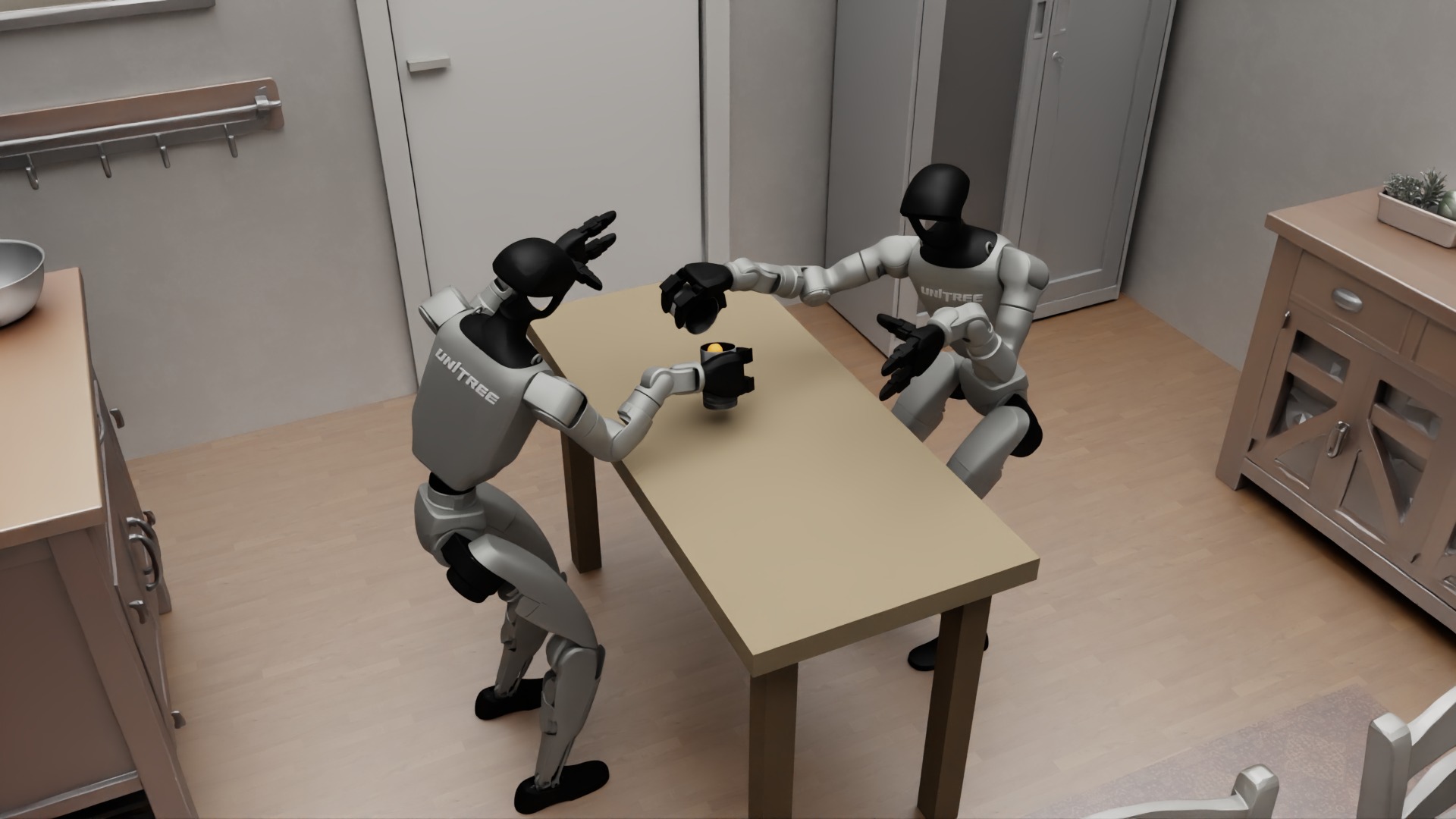}}\\[6pt]
  \taskcell{(c) FrameHang}{%
    \taskimg{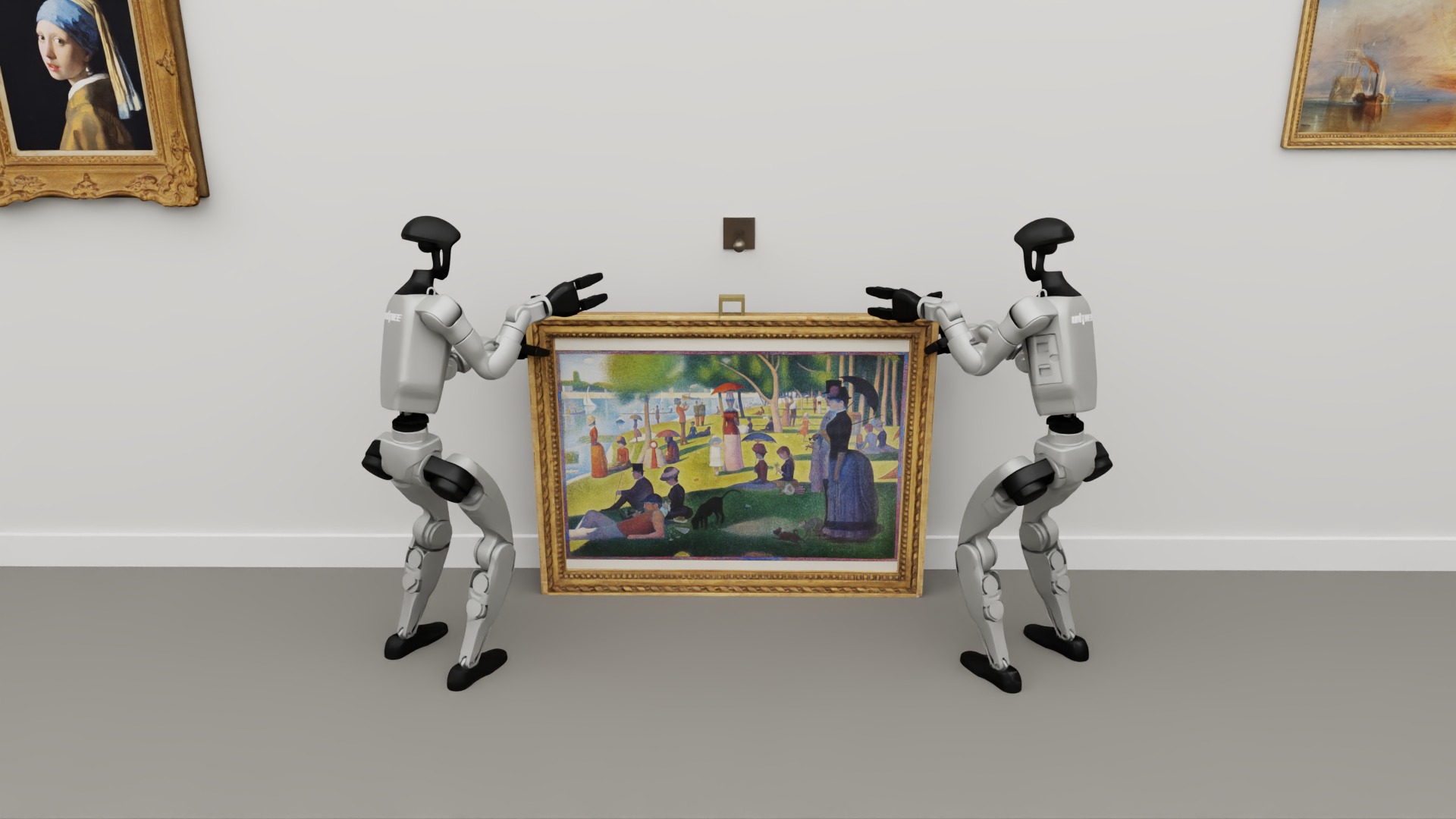}\hspace{\taskimgsep}%
    \taskimg{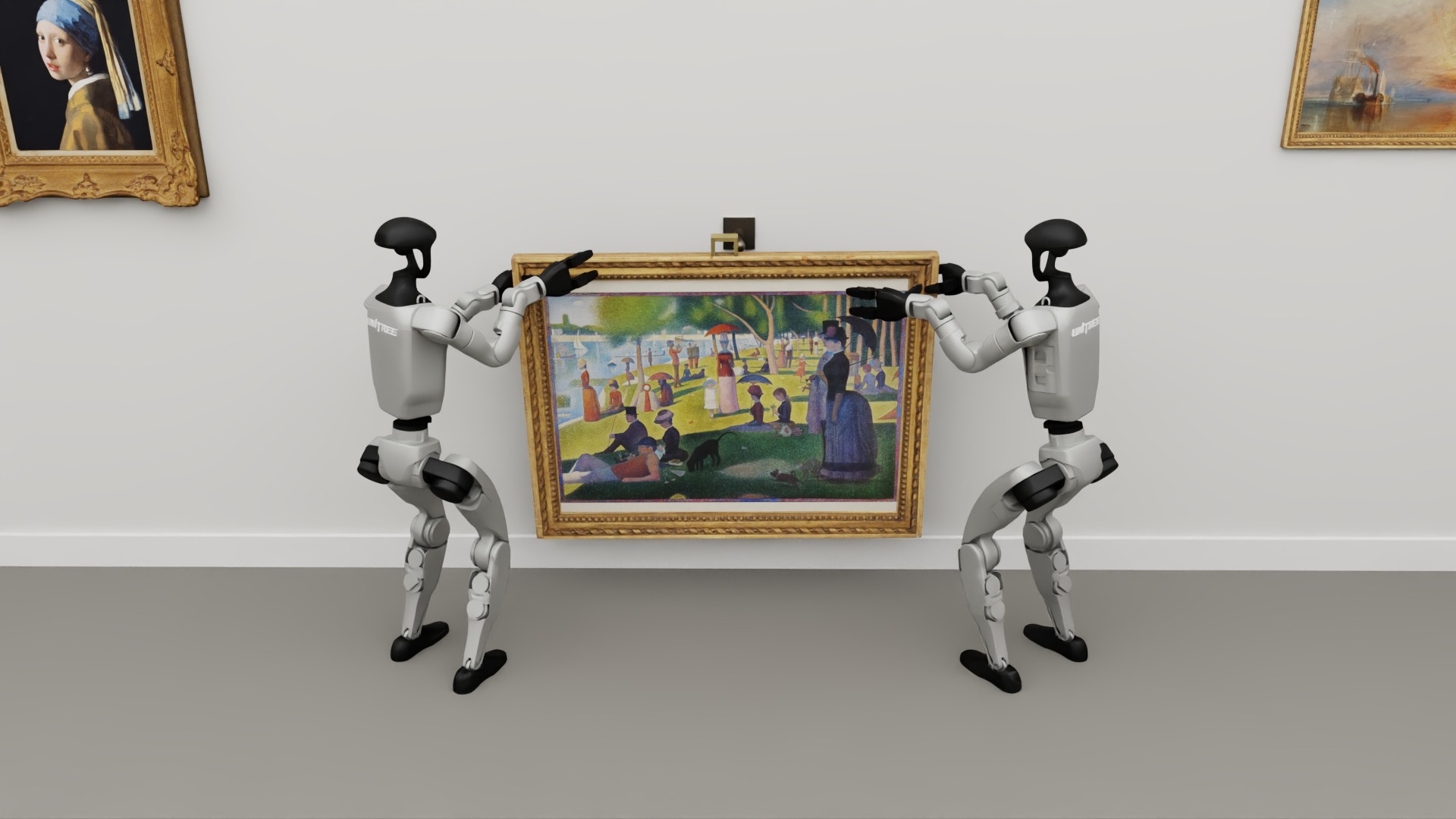}\hspace{\taskimgsep}%
    \taskimg{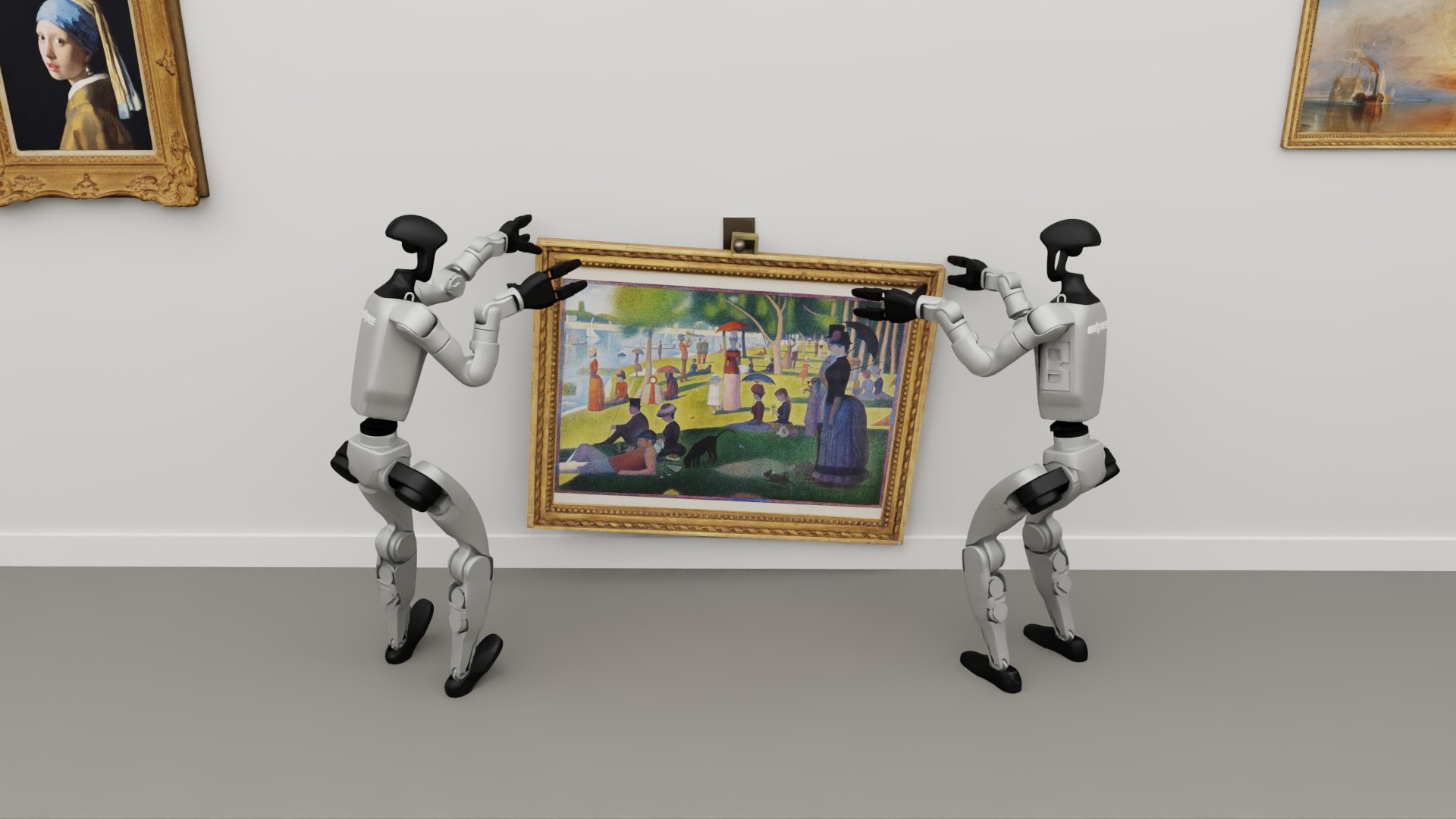}}\\[6pt]
  \taskcell{(d) CoPouring}{%
    \taskimg{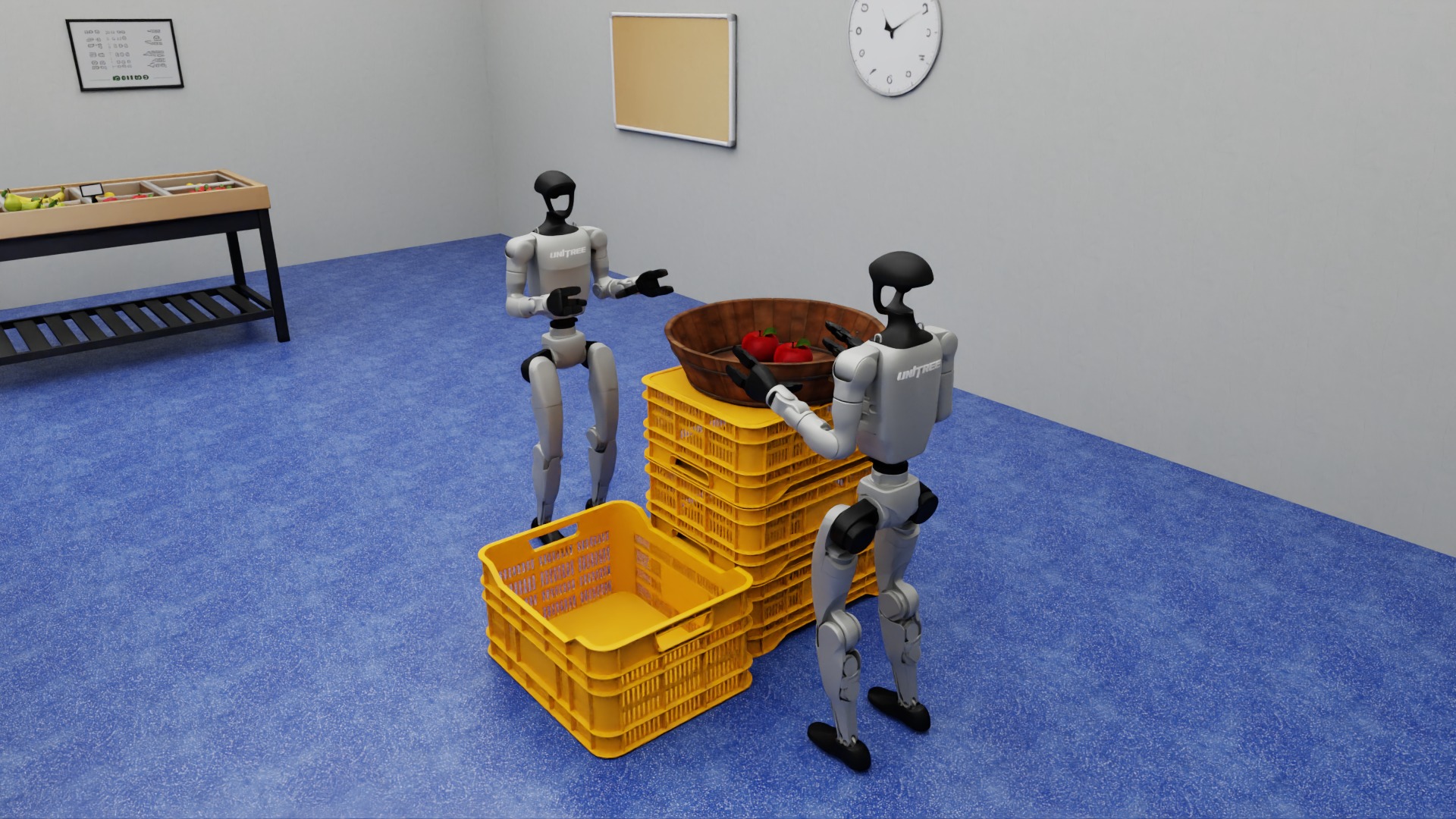}\hspace{\taskimgsep}%
    \taskimg{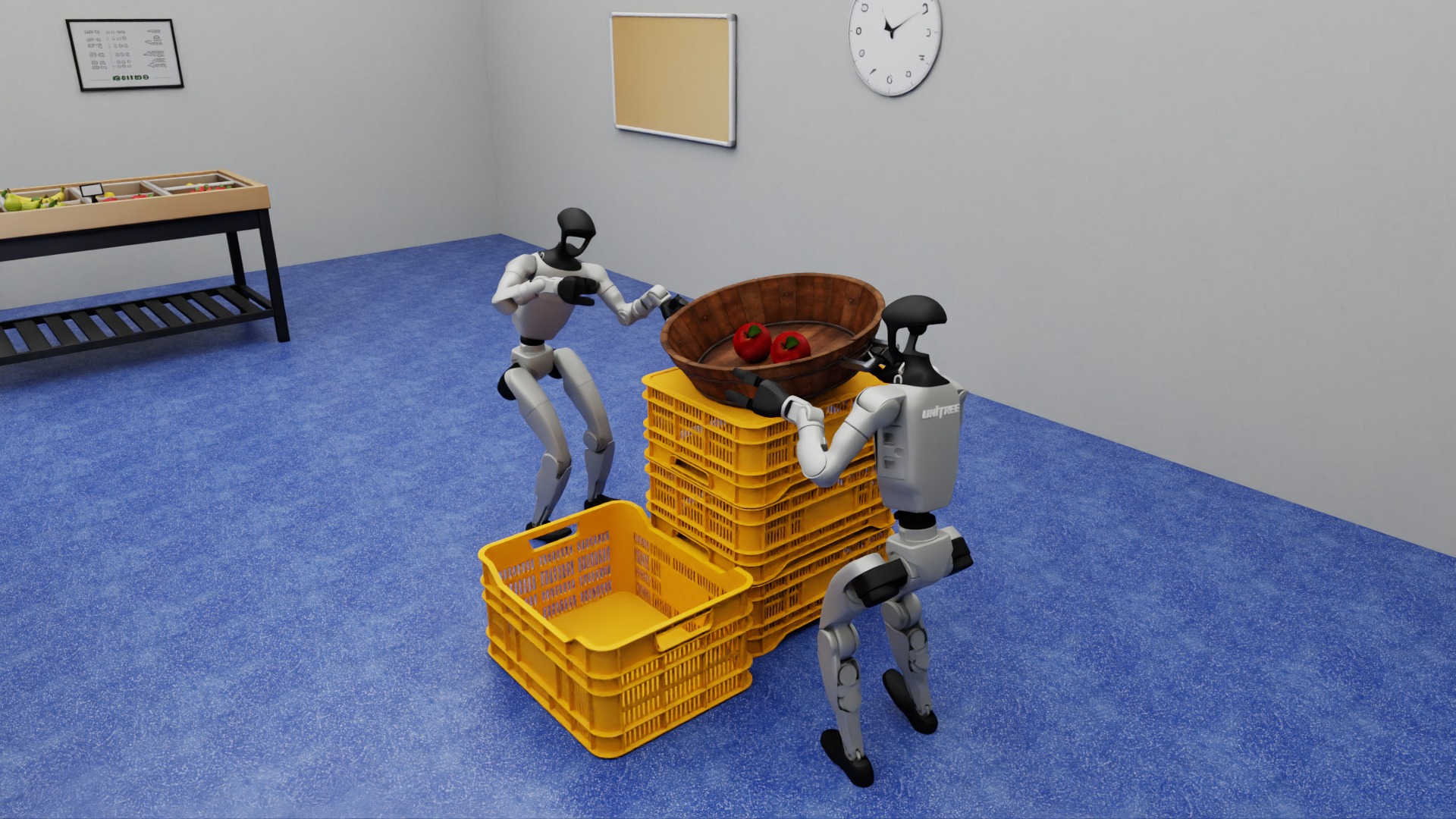}\hspace{\taskimgsep}%
    \taskimg{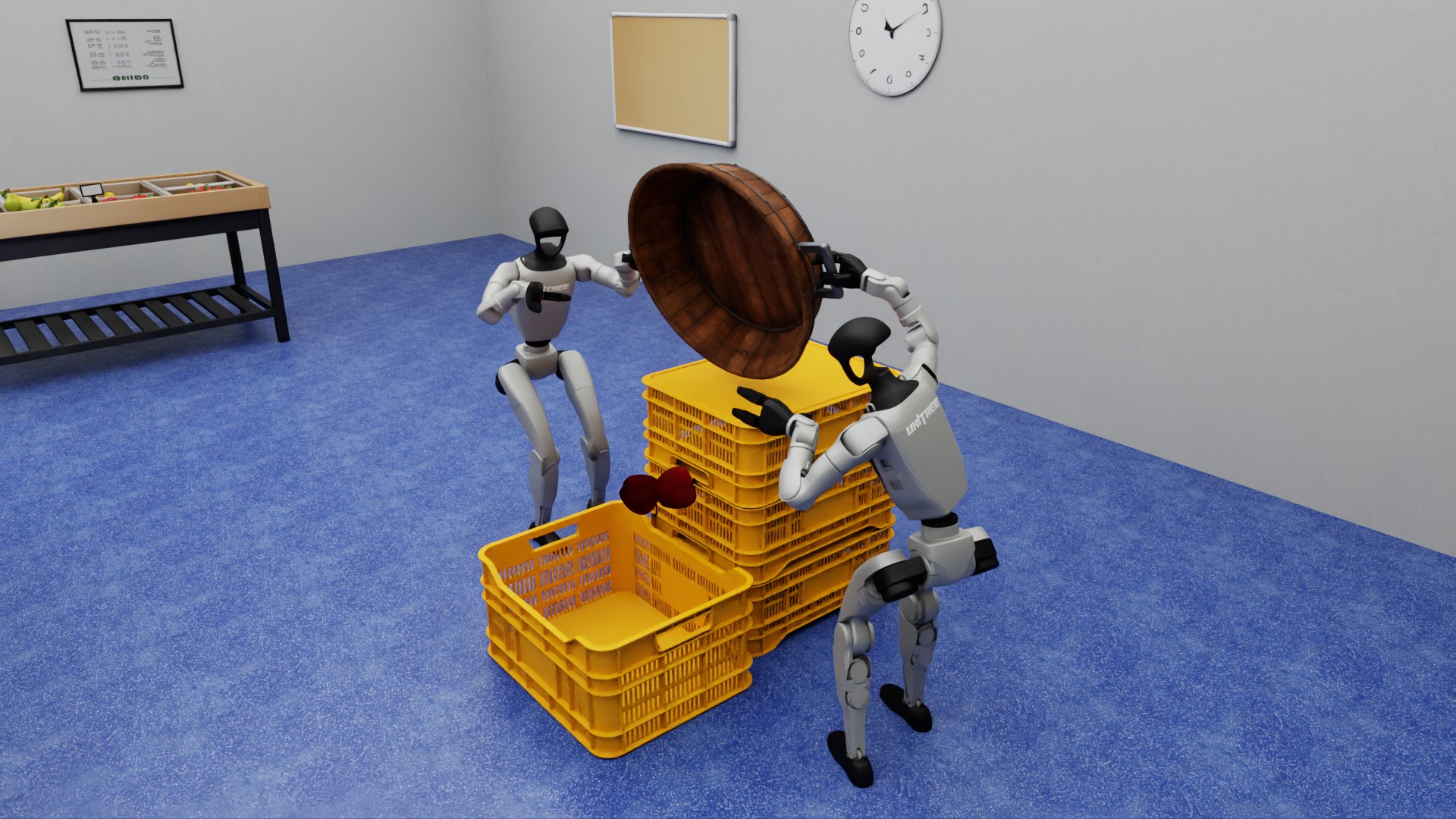}}
  \caption{\textbf{Key frames of the \bench tasks with two humanoids and low base movement.} Each row shows three key frames of one demonstration, ordered from left to right. (a)--(b) show tasks with low physical coupling, and (c)--(d) show tasks with high physical coupling.
  \textbf{(a) Handover:} one humanoid hands a bottle to its partner, which places it on the target.
  \textbf{(b) Pouring:} one humanoid pours a marble from a small cup into a glass held by its partner.
  \textbf{(c) FrameHang:} two humanoids jointly lift a large picture frame and hang it on a wall-mounted peg.
  \textbf{(d) CoPouring:} two humanoids lift a tub of apples together and tilt it to pour the apples into an empty crate.}
  \label{fig:gallery-two-low}
\end{figure}

\begin{figure}[p]
  \centering
  \setlength{\taskstripw}{0.84\linewidth}%
  \setlength{\taskimgsep}{0.015\linewidth}%
  \setlength{\taskframew}{\dimexpr(\taskstripw-2\taskimgsep)/3\relax}%
  \taskcell{(a) TrashCollection}{%
    \galleryimgtrim{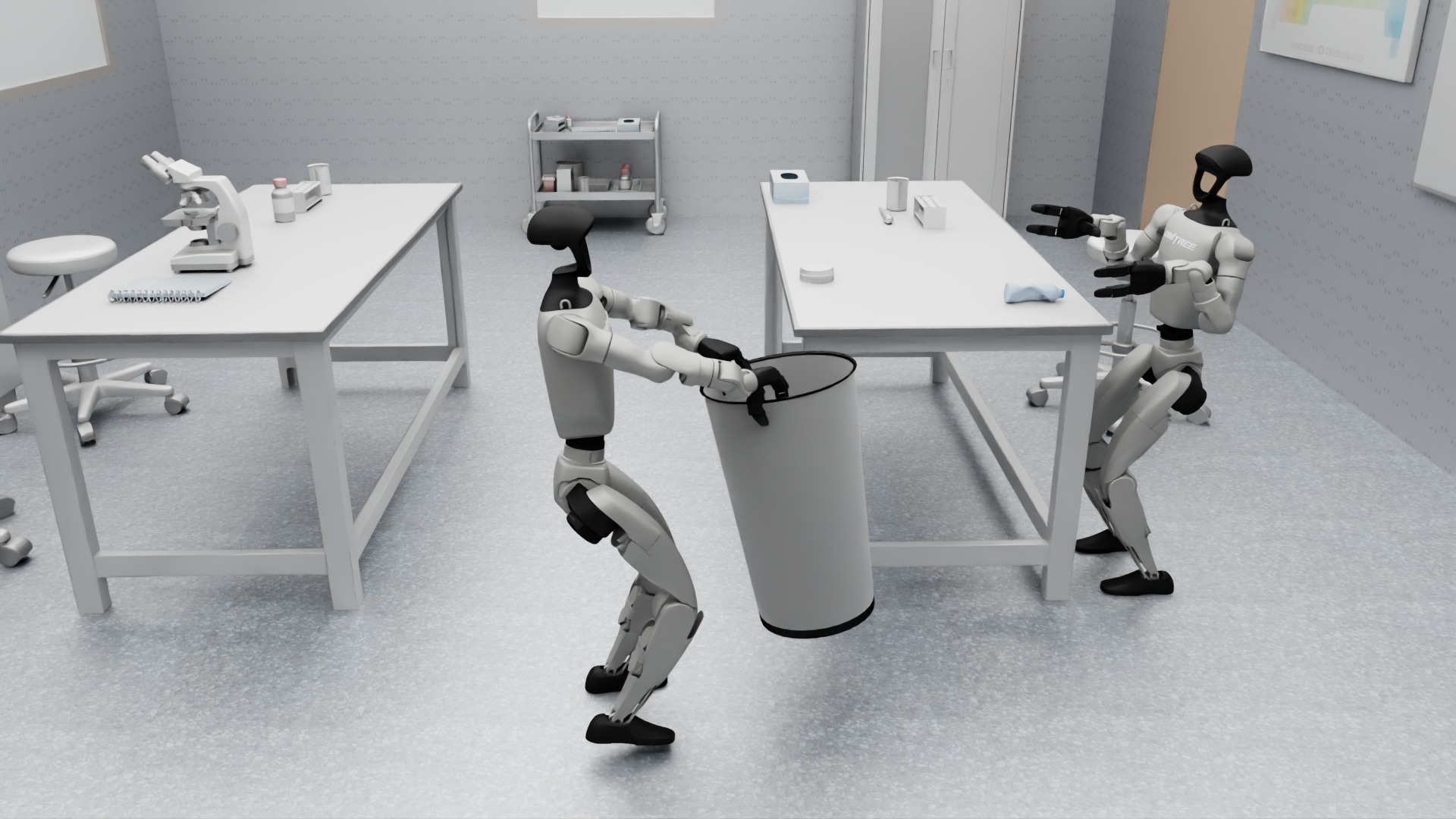}{0.322917}{0}{0.114583}{0}\hspace{\taskimgsep}%
    \galleryimgtrim{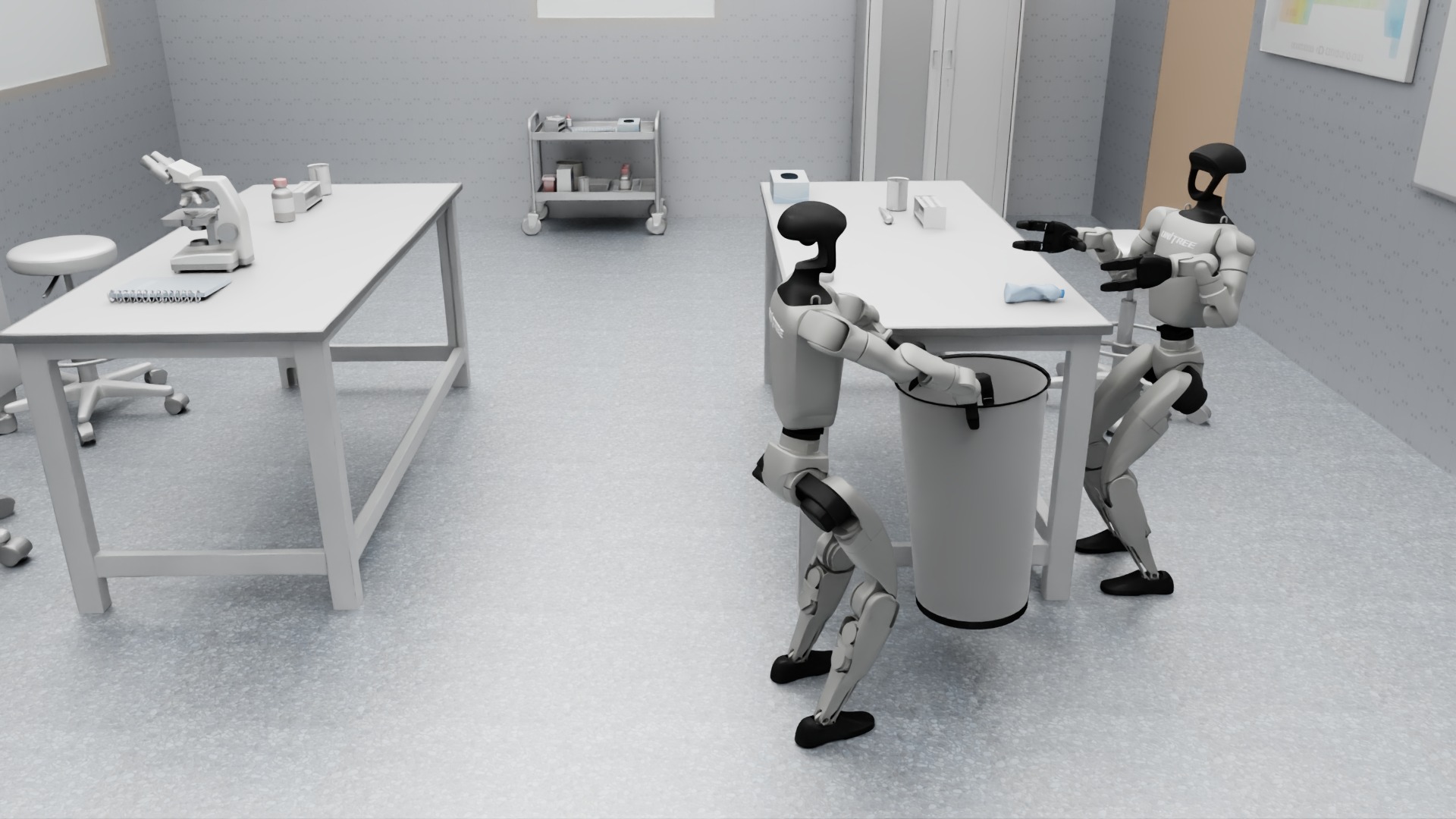}{0.416667}{0.055556}{0.083333}{0.055556}\hspace{\taskimgsep}%
    \galleryimgtrim{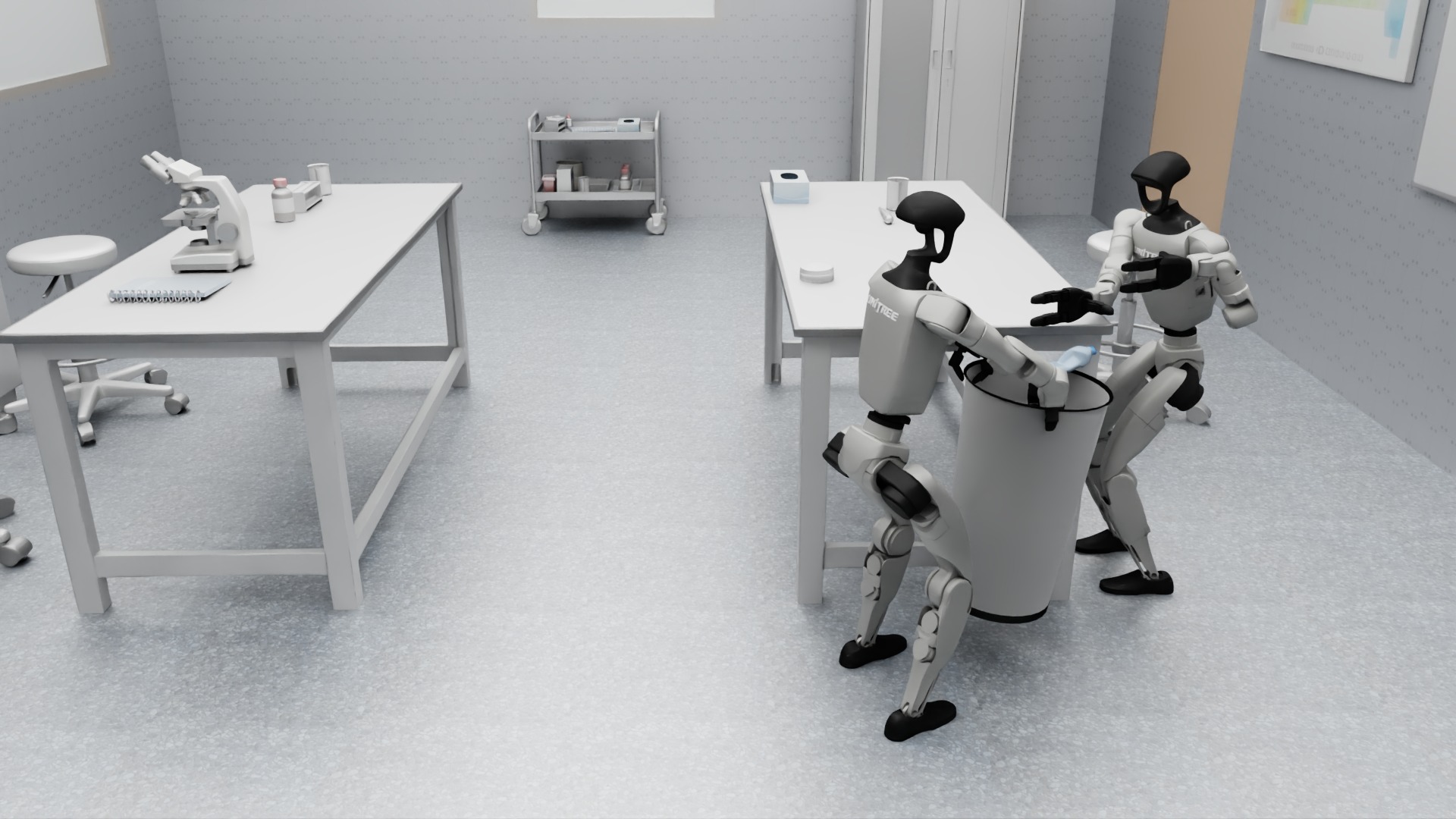}{0.416667}{0.055556}{0.083333}{0.055556}}\\[6pt]
  \taskcell{(b) CartService}{%
    \galleryimgtrim{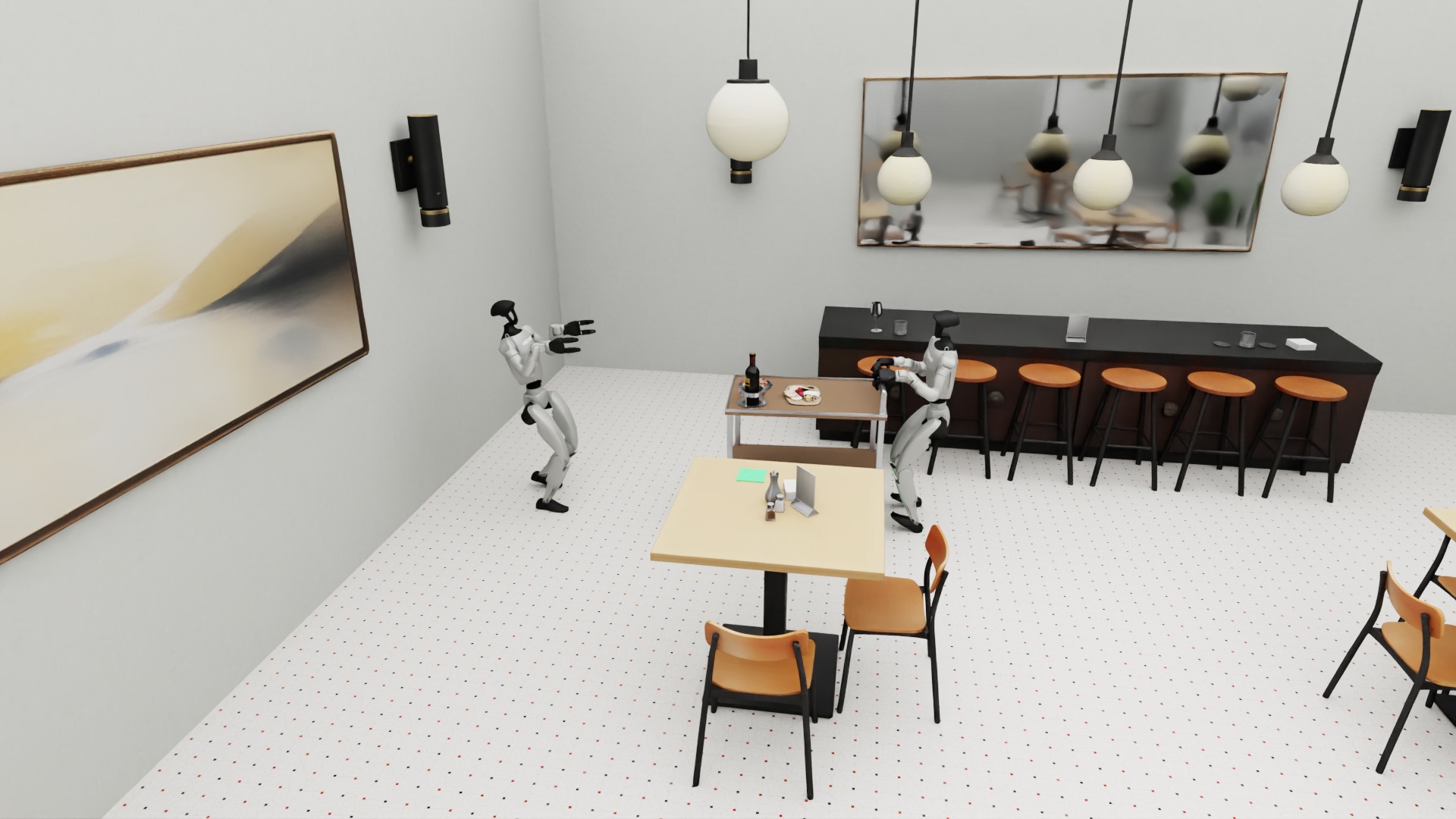}{0.338542}{0.157407}{0.317708}{0.231481}\hspace{\taskimgsep}%
    \galleryimgtrim{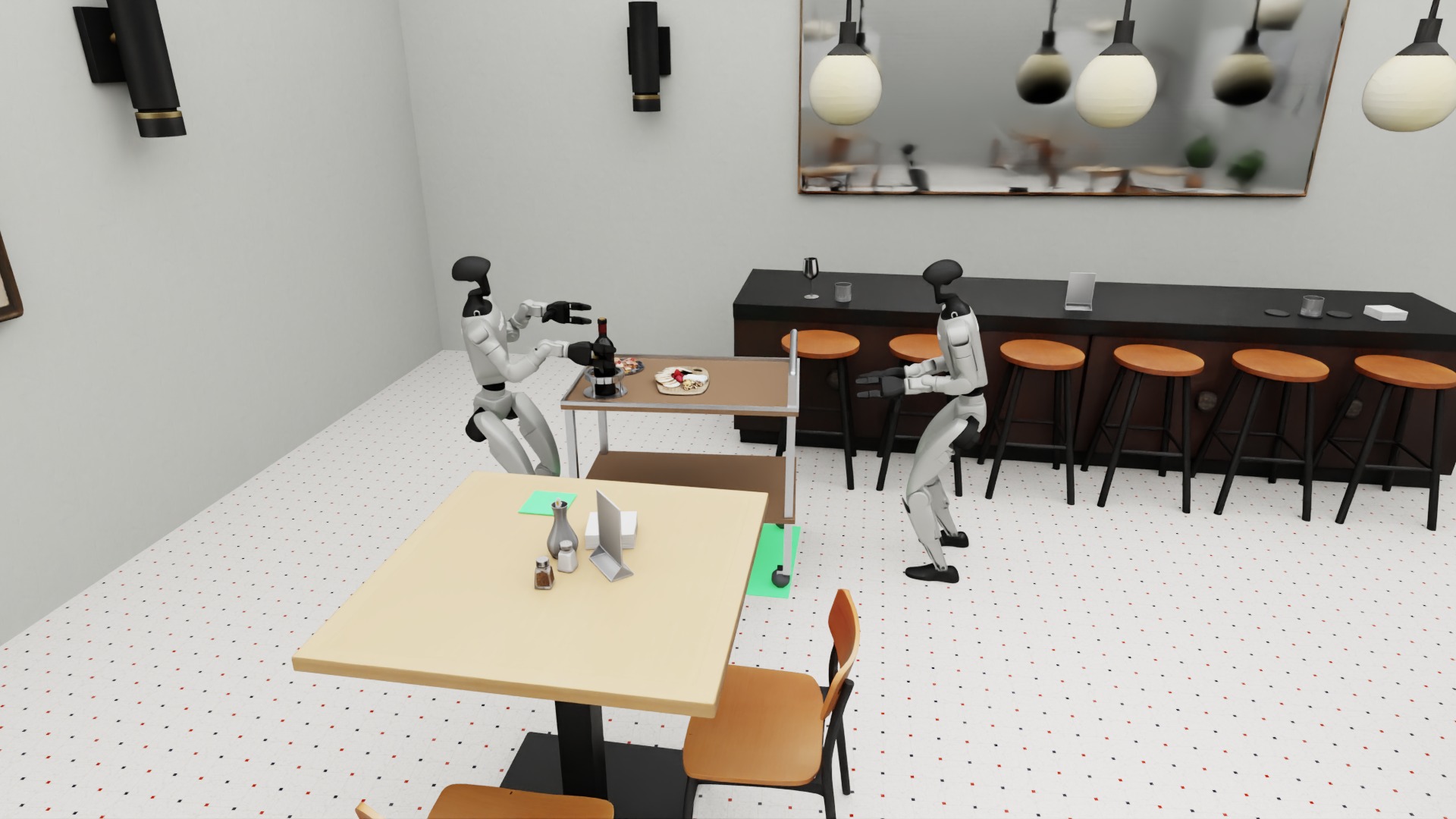}{0.244792}{0.046296}{0.244792}{0.046296}\hspace{\taskimgsep}%
    \galleryimgtrim{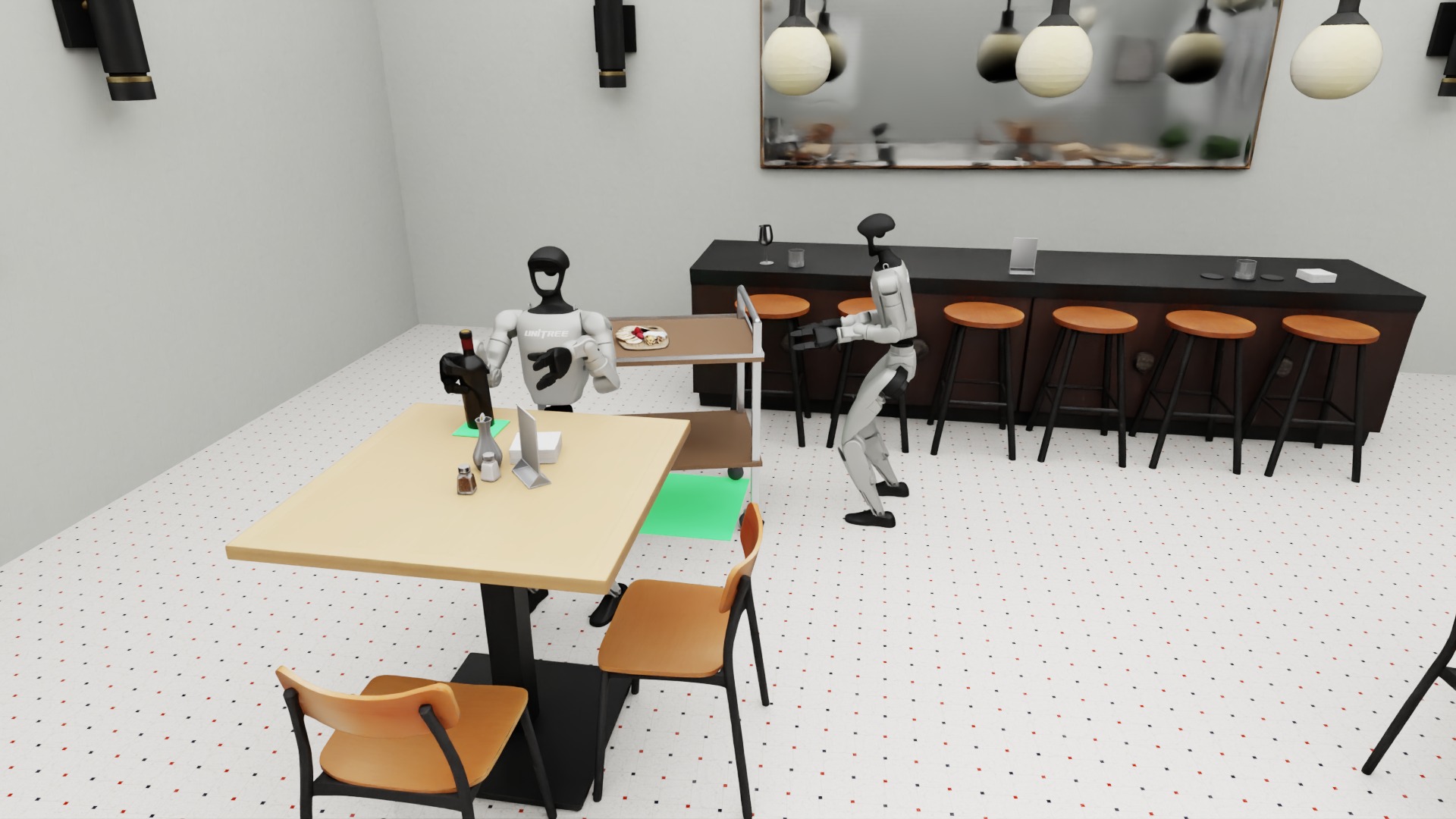}{0.218750}{0.046296}{0.270833}{0.046296}}\\[6pt]
  \taskcell{(c) CoCarry}{%
    \taskimg{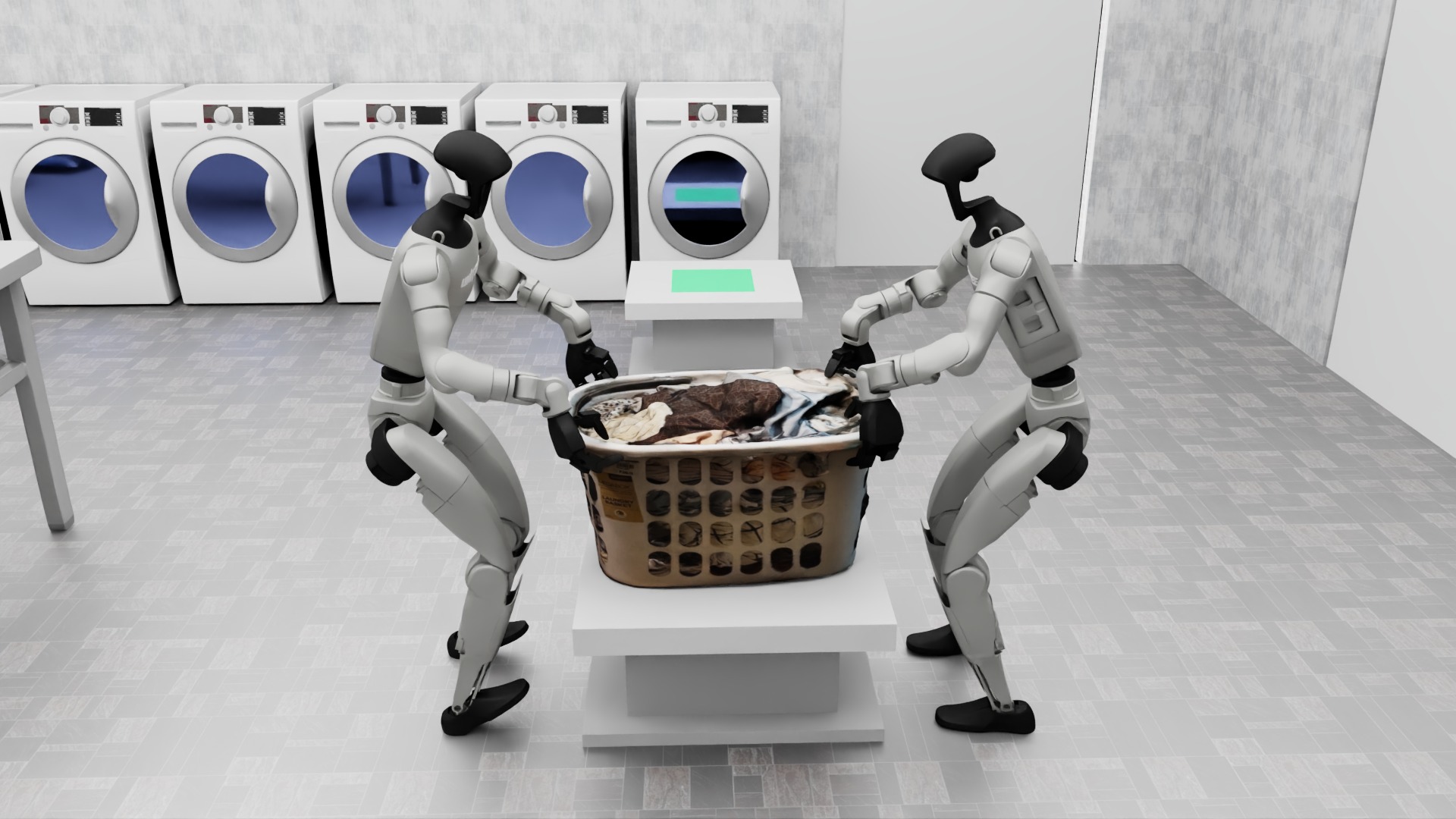}\hspace{\taskimgsep}%
    \taskimg{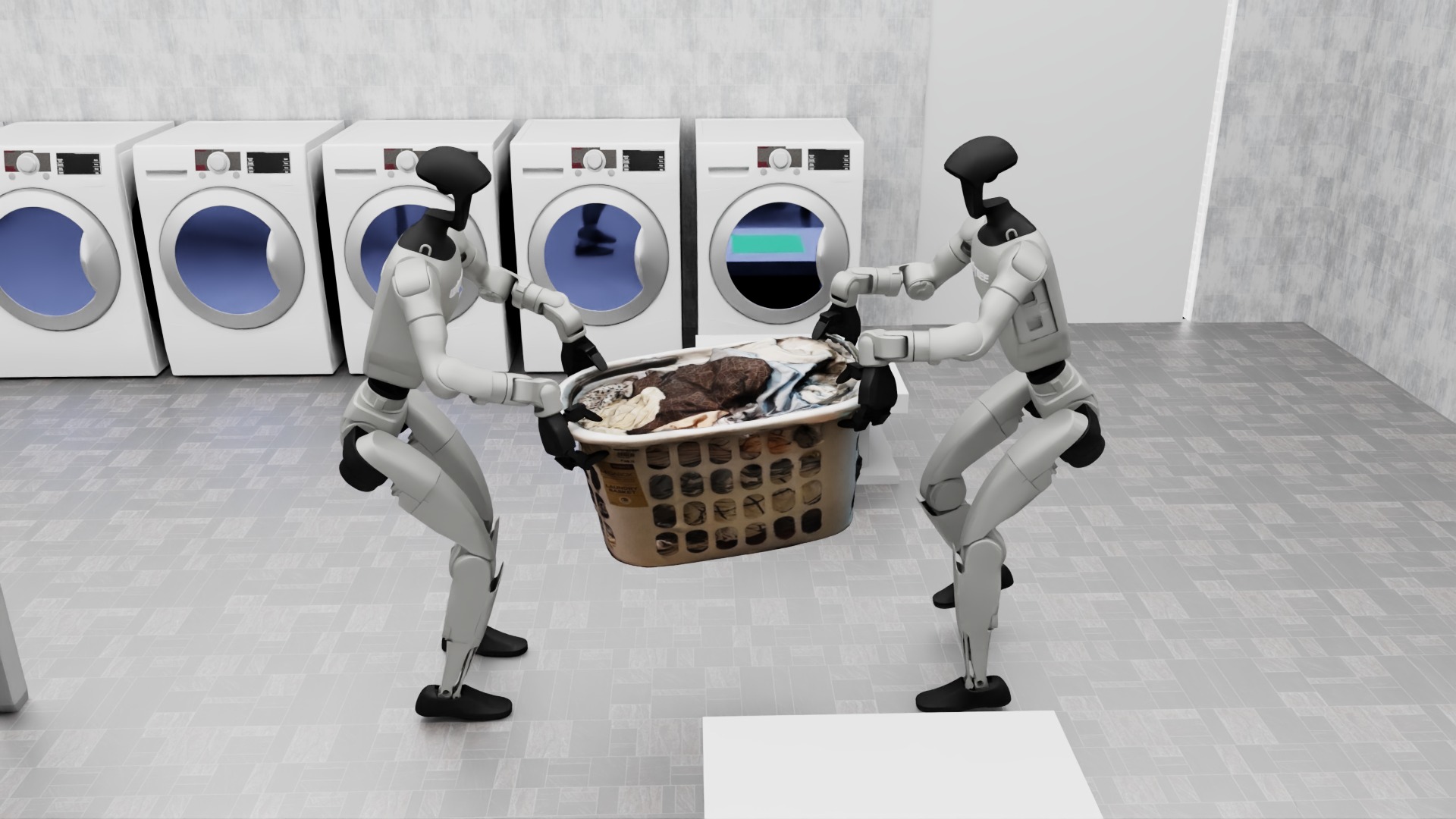}\hspace{\taskimgsep}%
    \taskimg{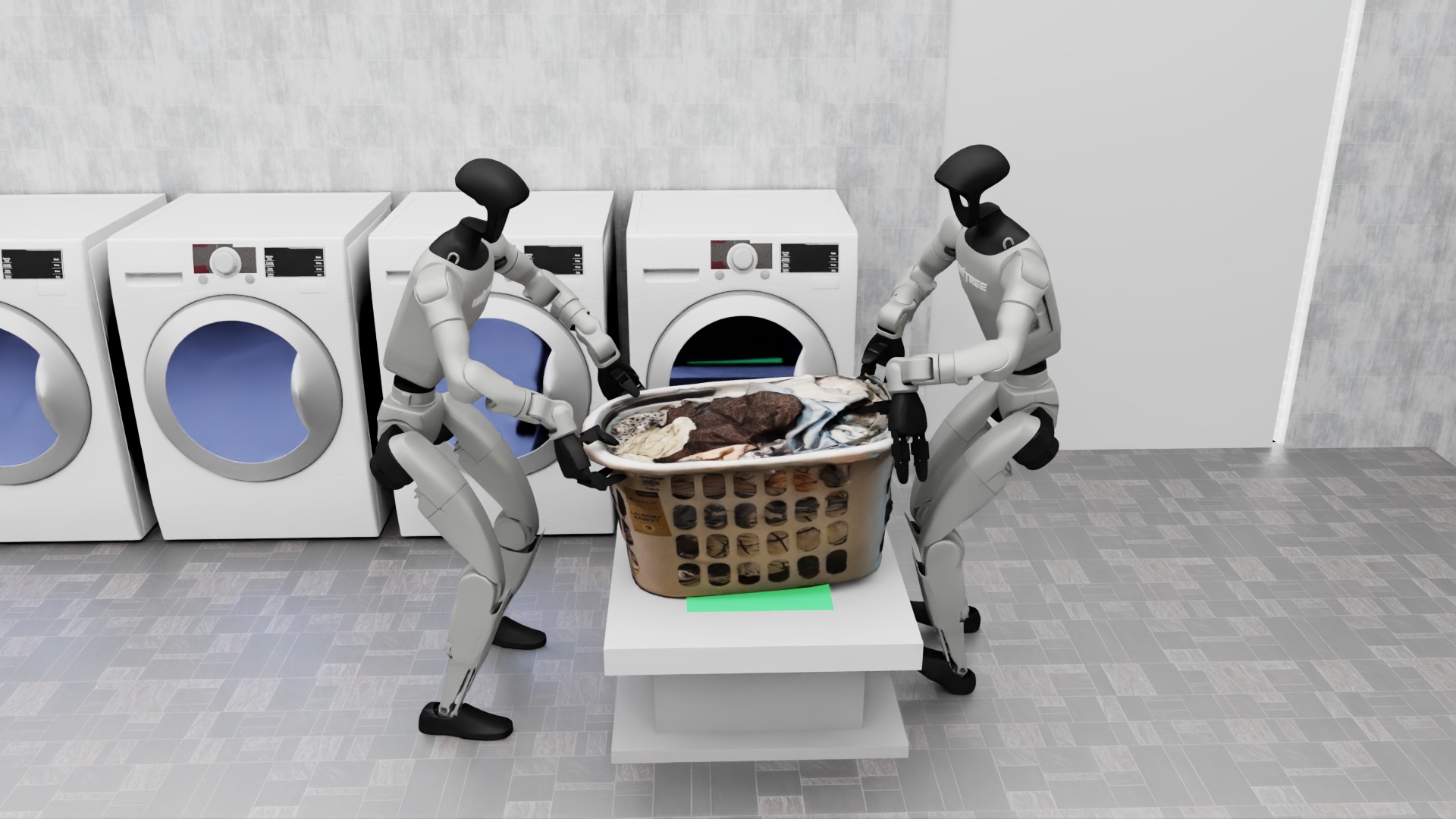}}\\[6pt]
  \taskcell{(d) TableAlign}{%
    \taskimg{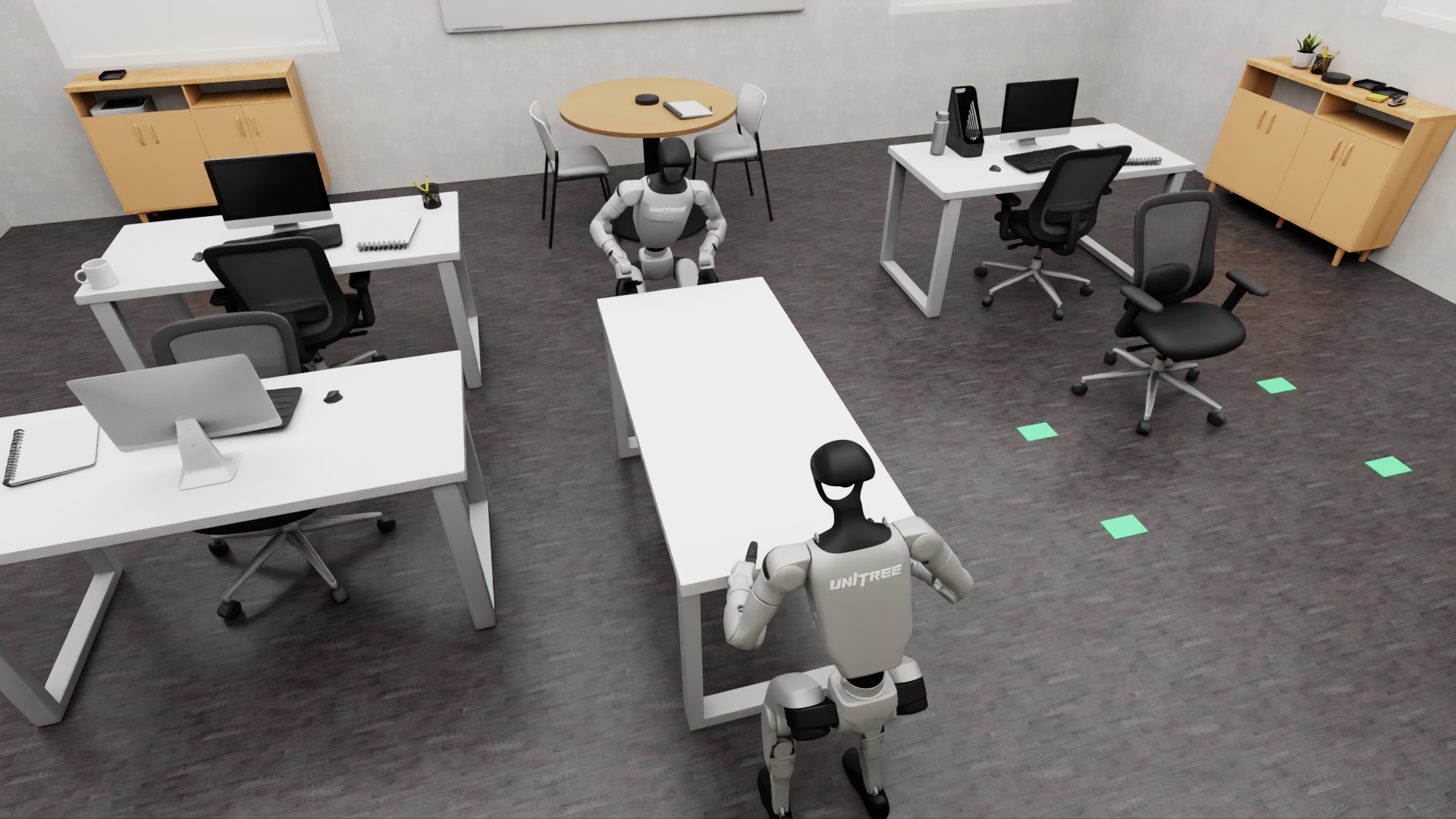}\hspace{\taskimgsep}%
    \galleryimgtrim{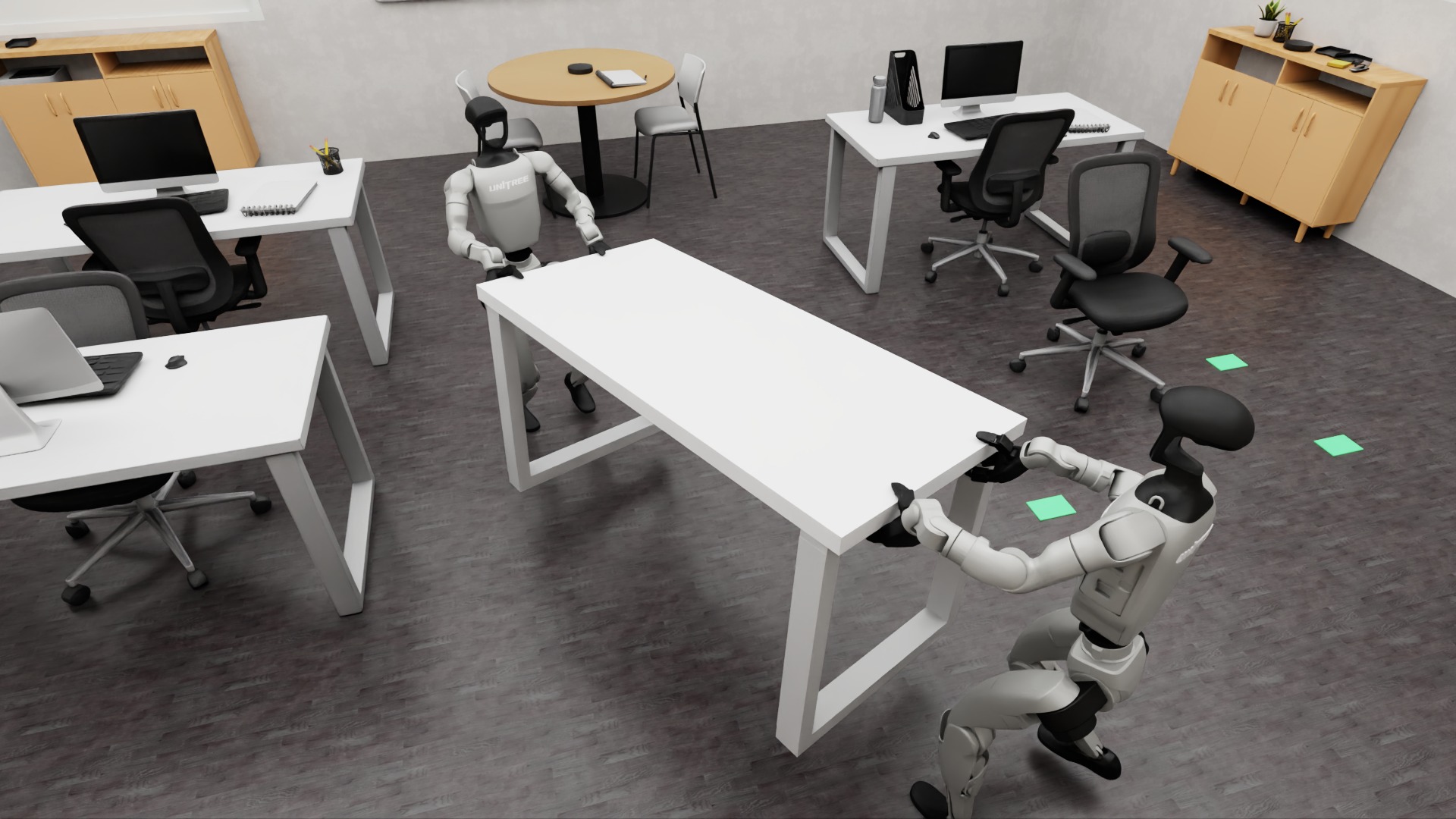}{0.281250}{0}{0.156250}{0}\hspace{\taskimgsep}%
    \galleryimgtrim{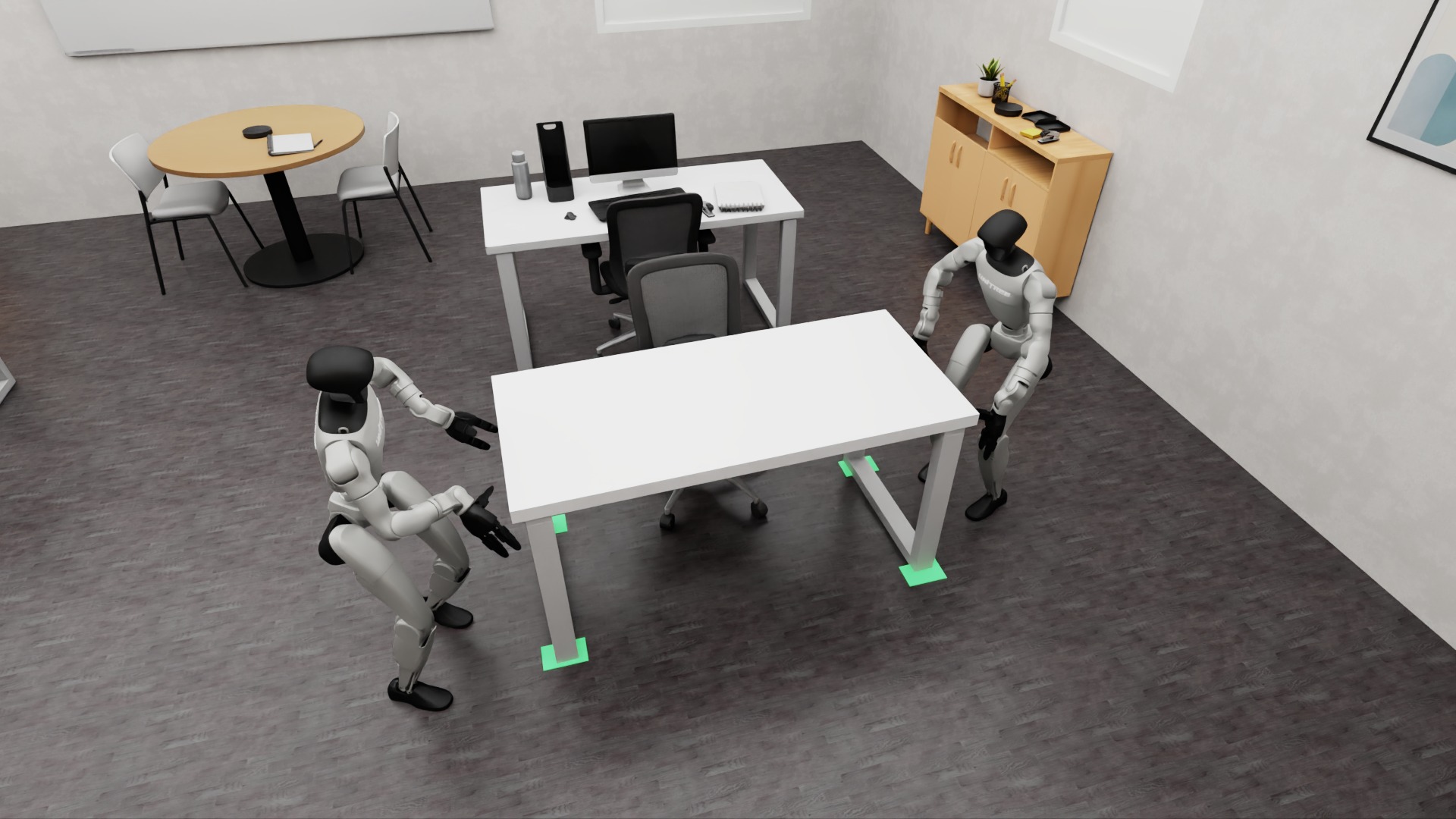}{0.177083}{0}{0.260417}{0}}
  \caption{\textbf{Key frames of the \bench tasks with two humanoids and high base movement.} Each row shows three key frames of one demonstration, ordered from left to right. (a)--(b) show tasks with low physical coupling, and (c)--(d) show tasks with high physical coupling.
  \textbf{(a) TrashCollection:} one humanoid carries a bin to the end of a table, where its partner sweeps a bottle into it.
  \textbf{(b) CartService:} one humanoid pushes a trolley to a table, and its partner takes a wine bottle from the trolley and places it on the table.
  \textbf{(c) CoCarry:} two humanoids jointly carry a laundry basket to a shelf while keeping it balanced.
  \textbf{(d) TableAlign:} two humanoids carry a table across an office, rotate it during transport, and place it aligned with a chair.}
  \label{fig:gallery-two-high}
\end{figure}

\begin{figure}[!t]
  \centering
  \setlength{\taskstripw}{0.84\linewidth}%
  \setlength{\taskimgsep}{0.015\linewidth}%
  \setlength{\taskframew}{\dimexpr(\taskstripw-2\taskimgsep)/3\relax}%
  \renewcommand{\taskcropside}{0}%
  \taskcell{(a) MoveHouse}{\taskstrip{figures/task_setup_figures/three_humanoids}%
    {movehouse_rollout_1}{movehouse_rollout_2}{movehouse_rollout_3}}\\[6pt]
  \taskcell{(b) BigTable}{\taskstrip{figures/task_setup_figures/three_humanoids}%
    {bigtable_rollout_1}{bigtable_rollout_2}{bigtable_rollout_3}}
  \caption{\textbf{Key frames of the \bench tasks with three humanoids.} Each row shows three key frames of one demonstration, ordered from left to right.
  \textbf{(a) MoveHouse:} one humanoid opens a door while the other two carry a table through the doorway and place it at the target location.
  \textbf{(b) BigTable:} all three humanoids jointly lift a large round table, carry it to the target location, and set it down together.}
  \label{fig:gallery-three}
\end{figure}

%% file: tables/phasewise-tables.tex
\section{Phase-Wise Results}
\label{app:phasewise}

We report cumulative success rates for the three phases $p_1 \rightarrow p_2 \rightarrow p_3$, defined per task in Appendix~\ref{app:task-phases}.
Tables~\ref{tab:app-phase-lowloco} and~\ref{tab:app-phase-highloco} cover tasks with two humanoids, split by base movement, and Table~\ref{tab:app-phase-3agent} covers tasks with three humanoids.
Blue and green family bands denote per-task and multi-task training, respectively.
For the tasks with two humanoids, the phase headings are colored as in Table~\ref{tab:app-task-phases}: \textcolor{collabphase}{\textbf{green}} for collaborative and \textcolor{noncollabphase}{\textbf{blue}} for non-collaborative phases.

\begin{table}[!ht]
  \caption{\textbf{Phase-wise success rates (\%) on tasks with two humanoids and low base movement.}
  Phases are defined in Appendix~\ref{app:tasks}; $p_3$ is full-task success.}
  \label{tab:app-phase-lowloco}
  \centering
  \footnotesize
  \setlength{\tabcolsep}{2.6pt}
  \renewcommand{\arraystretch}{1.05}
  \begin{tabular*}{\linewidth}{@{\extracolsep{\fill}}l*{12}{wc{17.5pt}}@{}}
    \toprule
    \textbf{Method} & \multicolumn{3}{c}{\textbf{Handover}} & \multicolumn{3}{c}{\textbf{Pouring}} & \multicolumn{3}{c}{\textbf{FrameHang}} & \multicolumn{3}{c}{\textbf{CoPouring}} \\
    \cmidrule(lr){2-4}\cmidrule(lr){5-7}\cmidrule(lr){8-10}\cmidrule(lr){11-13}
      & \noncollabphase{$p_1$} & \collabphase{$p_2$} & \noncollabphase{$p_3$} & \noncollabphase{$p_1$} & \noncollabphase{$p_2$} & \collabphase{$p_3$} & \noncollabphase{$p_1$} & \collabphase{$p_2$} & \collabphase{$p_3$} & \collabphase{$p_1$} & \collabphase{$p_2$} & \collabphase{$p_3$} \\
    \midrule
    \famrow{13}{Standard IL policies}
    ACT              & 100 & 98 & 65 & 95 & 93 & 1 & 100 & 100 & 44 & 33 & 32 & 9 \\
    DP               & 87 & 64 & 30 & 38 & 38 & 0 & 76 & 73 & 26 & 21 & 18 & 7 \\
    \addlinespace[2pt]
    \famrow{13}{Multi-agent IL policies}
    LatentToM        & 77 & 23 & 5 & 48 & 41 & 0 & 58 & 24 & 2 & 13 & 10 & 1 \\
    GauDP            & 92 & 85 & 52 & 29 & 28 & 0 & 81 & 76 & 29 & 0 & 0 & 0 \\
    \addlinespace[2pt]
    \famrowmt{13}{Vision-Language-Action (VLA) models}
    $\pi_{0.5}$      & 92 & 12 & 2 & 58 & 54 & 0 & 91 & 86 & 27 & 1 & 1 & 0 \\
    GR00T N1.7       & 95 & 77 & 27 & 63 & 59 & 0 & 94 & 90 & 44 & 2 & 2 & 0 \\
    $\Psi_0$         & 65 & 8 & 0 & 24 & 18 & 0 & 62 & 53 & 5 & 1 & 0 & 0 \\
    \addlinespace[2pt]
    \famrowmt{13}{World Action Models (WAMs)}
    Fast-WAM         & 98 & 62 & 2 & 86 & 84 & 7 & 97 & 94 & 28 & 28 & 25 & 11 \\
    \bottomrule
  \end{tabular*}
\end{table}

\begin{table}[!ht]
  \caption{\textbf{Phase-wise success rates (\%) on tasks with two humanoids and high base movement.}
  Phases are defined in Appendix~\ref{app:tasks}; $p_3$ is full-task success.}
  \label{tab:app-phase-highloco}
  \centering
  \footnotesize
  \setlength{\tabcolsep}{2.6pt}
  \renewcommand{\arraystretch}{1.05}
  \begin{tabular*}{\linewidth}{@{\extracolsep{\fill}}l*{12}{wc{17.5pt}}@{}}
    \toprule
    \textbf{Method} & \multicolumn{3}{c}{\textbf{TrashCollection}} & \multicolumn{3}{c}{\textbf{CartService}} & \multicolumn{3}{c}{\textbf{CoCarry}} & \multicolumn{3}{c}{\textbf{TableAlign}} \\
    \cmidrule(lr){2-4}\cmidrule(lr){5-7}\cmidrule(lr){8-10}\cmidrule(lr){11-13}
      & \noncollabphase{$p_1$} & \noncollabphase{$p_2$} & \collabphase{$p_3$} & \noncollabphase{$p_1$} & \noncollabphase{$p_2$} & \noncollabphase{$p_3$} & \collabphase{$p_1$} & \collabphase{$p_2$} & \collabphase{$p_3$} & \collabphase{$p_1$} & \collabphase{$p_2$} & \collabphase{$p_3$} \\
    \midrule
    \famrow{13}{Standard IL policies}
    ACT              & 52 & 16 & 1 & 69 & 33 & 3 & 99 & 72 & 61 & 100 & 79 & 39 \\
    DP               & 32 & 4 & 1 & 66 & 9 & 3 & 96 & 31 & 12 & 93 & 34 & 4 \\
    \addlinespace[2pt]
    \famrow{13}{Multi-agent IL policies}
    LatentToM        & 30 & 5 & 0 & 52 & 6 & 0 & 98 & 44 & 31 & 96 & 23 & 3 \\
    GauDP            & 52 & 6 & 1 & 64 & 2 & 0 & 92 & 22 & 9 & 95 & 62 & 16 \\
    \addlinespace[2pt]
    \famrowmt{13}{Vision-Language-Action (VLA) models}
    $\pi_{0.5}$      & 56 & 10 & 0 & 60 & 34 & 4 & 73 & 4 & 1 & 75 & 6 & 2 \\
    GR00T N1.7       & 49 & 13 & 1 & 77 & 36 & 0 & 49 & 7 & 2 & 62 & 6 & 1 \\
    $\Psi_0$         & 55 & 18 & 3 & 77 & 2 & 0 & 82 & 0 & 0 & 77 & 0 & 0 \\
    \addlinespace[2pt]
    \famrowmt{13}{World Action Models (WAMs)}
    Fast-WAM         & 55 & 11 & 1 & 73 & 45 & 0 & 99 & 53 & 31 & 96 & 40 & 17 \\
    \bottomrule
  \end{tabular*}
\end{table}

\begin{table}[!ht]
  \caption{\textbf{Phase-wise success rates (\%) on tasks with three humanoids.}
  Phases are defined in Appendix~\ref{app:tasks}; $p_3$ is full-task success.}
  \label{tab:app-phase-3agent}
  \centering
  \footnotesize
  \setlength{\tabcolsep}{2.6pt}
  \renewcommand{\arraystretch}{1.05}
  \begin{tabular*}{\linewidth}{@{\extracolsep{\fill}}lcccccc@{\hspace{1.5em}}}
    \toprule
    \textbf{Method} & \multicolumn{3}{c}{\textbf{MoveHouse}} & \multicolumn{3}{c@{\hspace{1.5em}}}{\textbf{BigTable}} \\
    \cmidrule(lr){2-4}\cmidrule(lr{1.5em}){5-7}
      & $p_1$ & $p_2$ & $p_3$ & $p_1$ & $p_2$ & $p_3$ \\
    \midrule
    \famrow{7}{Standard IL policies}
    ACT              & 89 & 80 & 9 & 100 & 98 & 80 \\
    DP               & 73 & 57 & 4 & 98 & 50 & 14 \\
    \addlinespace[2pt]
    \famrow{7}{Multi-agent IL policies}
    LatentToM$^{\dagger}$ & \na & \na & \na & \na & \na & \na \\
    GauDP            & 38 & 30 & 2 & 100 & 83 & 54 \\
    \addlinespace[2pt]
    \famrowmt{7}{Vision-Language-Action (VLA) models}
    $\pi_{0.5}$      & 33 & 17 & 3 & 77 & 44 & 37 \\
    GR00T N1.7       & 40 & 22 & 0 & 36 & 9 & 5 \\
    $\Psi_0$         & 61 & 45 & 0 & 87 & 23 & 19 \\
    \addlinespace[2pt]
    \famrowmt{7}{World Action Models (WAMs)}
    Fast-WAM         & 95 & 80 & 20 & 97 & 94 & 69 \\
    \bottomrule
  \end{tabular*}
  \par\vspace{3pt}
  \begin{minipage}{\linewidth}
    \footnotesize\raggedright
    $^{\dagger}$ We evaluate LatentToM only on tasks with two robots, following
    the original paper's experimental setup.
  \end{minipage}
\end{table}